\documentclass[12pt]{article}

\usepackage{amssymb}
\usepackage{amsmath,amsthm,amssymb,amsopn,amsfonts,pdfpages,url}
\usepackage[linesnumbered,ruled,vlined]{algorithm2e}
\usepackage{graphics,relsize}
\usepackage{graphicx}
\usepackage{multirow}
\usepackage{subfig}
\usepackage{bbm}
\usepackage{comment}
\usepackage{booktabs}
\usepackage[square,numbers]{natbib}
\newcommand{\grr}{\mathbbm{1}}
\newcommand{\tP}{\text{P}}

\newcommand{\twospace}{\renewcommand{\baselinestretch}{1.3}\normalsize}

\newcommand{\bmu}{\boldsymbol{\mu}}
\newcommand{\bsigma}{\boldsymbol{\sigma}}

\newcommand{\q}{\mathbb{Q}}
\newcommand{\e}{\mathbb{E}}
\newcommand{\br}{b_{\restriction S}}

\newtheorem{theorem}{Theorem}
\newtheorem{defn}[theorem]{Definition}

\twospace
\begin{document}

\title{Vertex nomination: The canonical sampling and the extended
spectral nomination schemes}
\maketitle
\author{ \noindent
Jordan Yoder$^1$,
Li Chen$^2$,
Henry Pao$^3$,
Eric Bridgeford$^4$,
Keith Levin$^5$,
Donniell E. Fishkind$^6$,
Carey Priebe$^6$,
Vince Lyzinski$^7$.\\
{\it 1: Jordan $\&$ Yoder LLC, 2: Intel Labs, 3: Amazon.com, 4: Department of Biostatistics, Johns Hopkins University, 5: Department of Statistics, University of Michigan, 6: Department of Applied Mathematics and Statistics, Johns Hopkins University, 7: Department of Mathematics and Statistics, University of Massachusetts Amherst
}

\begin{abstract}  \vspace{-.1in}
Suppose that one particular block in a stochastic block model is of interest,
but block labels are only observed for a few of the vertices in the network.
Utilizing a graph realized from the model and the observed block labels,
the vertex nomination task is to order the vertices with unobserved block
labels into a ranked nomination list
with the goal of having an abundance of interesting vertices near the top of the list.
There are vertex nomination schemes in the literature, including
the optimally precise canonical nomination scheme~$\mathcal{L}^C$
and the consistent spectral partitioning nomination scheme~$\mathcal{L}^P$.
While the canonical nomination scheme $\mathcal{L}^C$ is provably optimally precise, it is
computationally intractable, being impractical to implement even on modestly
sized graphs.\\ \indent
With this in mind, an approximation of the canonical
scheme---denoted the {\it canonical sampling nomination scheme} $\mathcal{L}^{CS}$---is introduced; $\mathcal{L}^{CS}$ relies on a scalable, Markov chain Monte Carlo-based approximation of $\mathcal{L}^{C}$, and converges to
$\mathcal{L}^{C}$ as the amount of sampling goes to infinity. The spectral partitioning nomination scheme is also extended to the
{\it extended spectral partitioning nomination scheme}, $\mathcal{L}^{EP}$, which
introduces a novel semisupervised clustering framework to improve
upon the precision of $\mathcal{L}^P$. Real-data and simulation experiments
are employed to illustrate the precision of these vertex nomination schemes,
as well as their empirical computational complexity.
\end{abstract}

\section{Introduction}
\label{S:intro}

Network data often exhibits underlying community structure, and there is a vast literature devoted
to uncovering communities in complex networks;
see, for example, \cite{newman2006modularity,von2007tutorial,rohe2011spectral,sussman12:_univer}.
In many applications, one community in the network is of particular interest to the researcher.
For example, in neuroscience connectomics, researchers might want to identify the region of the
brain responsible for a particular neurological function; in a social network, a marketing company
might want to find a group of users with similar interests; in an internet hyperlink network, a
journalist might want to find blogs with a certain political leaning or subject matter.
If we are given a few vertices known to be from from the community of interest, and perhaps a
few vertices known to not be from the community of interest, the task of {\it vertex nomination}
is to order the remaining vertices in the network into a {\it nomination list}, with the aim of
having a concentration of vertices from the community of interest at the top of the list;
for alternate formulations of the vertex nomination problem, see \cite{HP,LKP}.

In \cite{fishkind2015vertex}, three novel vertex nomination schemes were introduced:
the canonical vertex nomination scheme $\mathcal{L}^C$,
the likelihood maximization vertex nomination scheme $\mathcal{L}^{ML}$,
and the spectral partitioning vertex nomination scheme $\mathcal{L}^P$.
Under mild model assumptions, the canonical vertex nomination scheme $\mathcal{L}^C$---which
is the vertex nomination analogue of the Bayes' classifier---was proven to be the optimal vertex
nomination scheme according to a mean average precision metric (see Definition \ref{defn:MAP}).
Unfortunately, $\mathcal{L}^C$ is not practical to implement on graphs with more than a few tens of vertices.
The likelihood maximization vertex nomination scheme $\mathcal{L}^{ML}$
utilizes novel graph matching machinery, and is shown to be highly effective on both simulated and real data sets.
However, $\mathcal{L}^{ML}$ is not practical to implement on graphs with more than a few thousand vertices.
The spectral partitioning vertex nomination scheme $\mathcal{L}^P$
is less effective than the canonical and the likelihood maximization vertex nomination schemes on the
small and moderately sized networks
where the canonical and the likelihood maximization vertex nomination schemes can respectively be implemented in practice.
Nonetheless, the spectral partitioning vertex nomination scheme has the significant advantage of being
practical to implement on graphs with up to tens of millions of vertices.

\subsection{ Extending $\mathcal{L}^C$ and $\mathcal{L}^P$}
In this paper we present extensions of the $\mathcal{L}^C$ and $\mathcal{L}^P$ vertex nomination schemes.
Our extension of the canonical vertex nomination scheme $\mathcal{L}^C$, which we shall call the
{\it canonical sampling vertex nomination scheme} and denote it as $\mathcal{L}^{CS}$,
is an approximation of $\mathcal{L}^C$ that can be practically computed for graphs with hundreds of
thousands of vertices, and our extension of the spectral partitioning vertex nomination scheme $\mathcal{L}^P$,
which we shall call the {\it extended spectral partitioning vertex
nomination scheme} and denote it as $\mathcal{L}^{EP}$, can be practically computed for graphs with
close to one hundred thousand vertices, with significantly increased effectiveness (i.e.~precision) over that of $\mathcal{L}^P$
when used on moderately sized networks.

While both
$\mathcal{L}^{CS}$ and $\mathcal{L}^{EP}$ are practical to implement
on very large graphs,
the former has the important theoretical advantage of
directly approximating the provably optimally precise vertex nomination scheme $\mathcal{L}^C$,
with this approximation getting better and better
when more and more sampling is used (and converging to $\mathcal{L}^C$ in this limit).
However, as with $\mathcal{L}^C$, the canonical sampling scheme can be held back by the
need to know/estimate the parameters of the underlying graph model before implementation.
While this may be impractical in settings where these estimates are infeasible, $\mathcal{L}^{CS}$
allows us to approximately compute optimal precision in a larger array of synthetic models, thereby
allowing us to better assess the performance of other, more feasibly implemented, procedures.
Indeed, given unlimited computational resources (for sampling purposes), when the model
parameters are known a priori or estimated to a suitable
precision, $\mathcal{L}^{CS}$ would be more effective than every vertex
nomination scheme other than $\mathcal{L}^C$.

In contrast, $\mathcal{L}^{EP}$ is implemented without needing to estimate the underlying
graph model parameters; indeed, including known parameter estimates into the
$\mathcal{L}^{EP}$ framework is nontrivial.
This can lead to superior performance of $\mathcal{L}^{EP}$ versus $\mathcal{L}^{CS}$,
especially in the setting where parameter estimates are necessarily highly variable.
Additionally, given equal computational resources (i.e., when limiting the sampling
allowed in $\mathcal{L}^{CS}$), $\mathcal{L}^{EP}$ is often more effective than $\mathcal{L}^{CS}$,
even when the model parameters are well estimated.

\begin{figure*}[t!]
\centering
\includegraphics[width=0.75\textwidth]{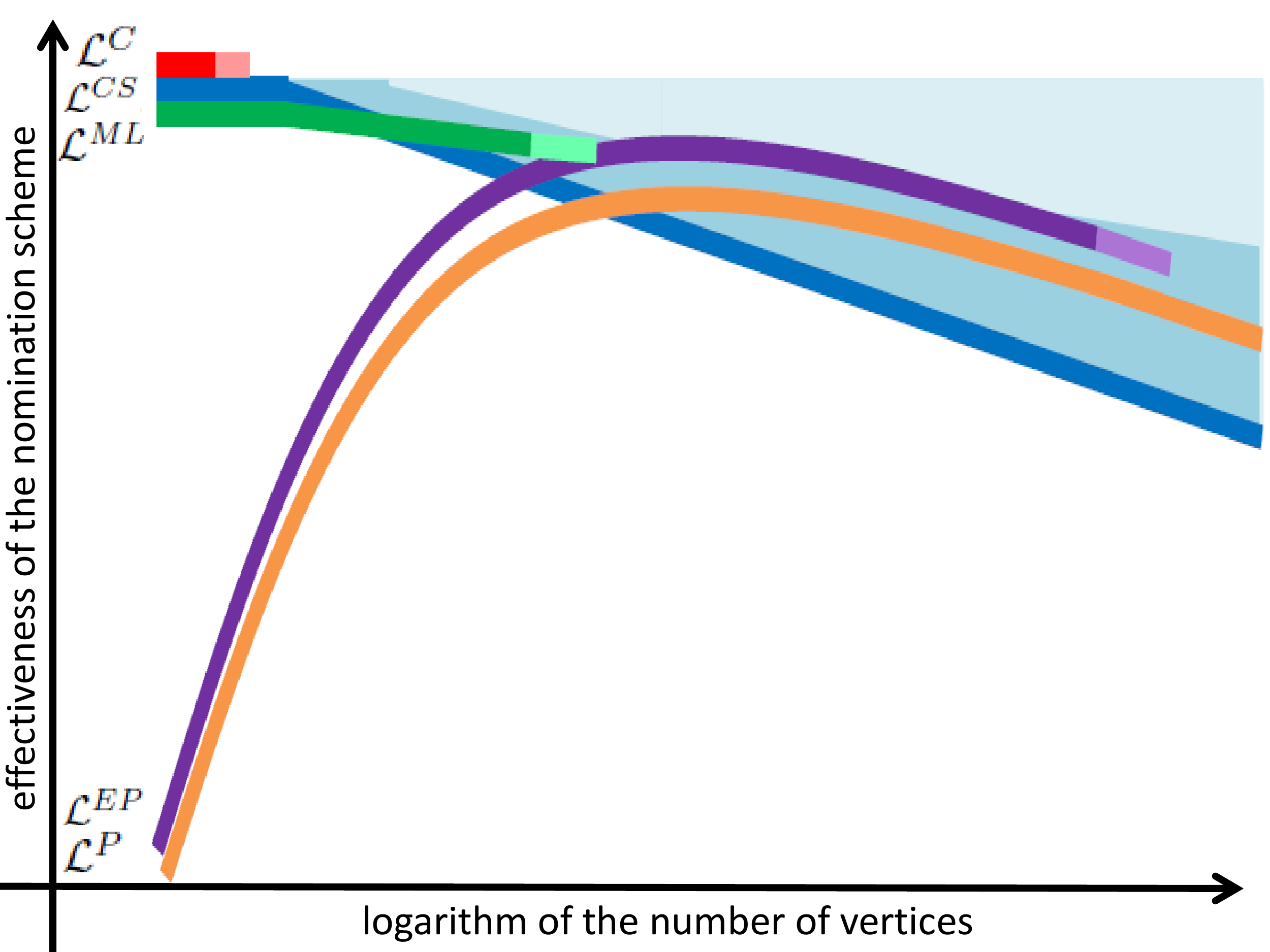}
\caption{A visual representation to summarize and compare the effectiveness (i.e.~precision) and
computational practicality of the vertex nomination schemes. This manuscript
introduces the canonical sampling vertex nomination scheme $\mathcal{L}^{CS}$ (blue)
as an extension of the canonical vertex nomination scheme $\mathcal{L}^C$ (red), and
introduces the extended spectral partitioning vertex nomination scheme
$\mathcal{L}^{EP}$ (purple) as a refinement of the spectral partitioning vertex nomination
scheme $\mathcal{L}^P$ (orange).}
\label{fig:art}
\end{figure*}

See Figure \ref{fig:art} for an informal visual representation that succinctly
compares the various vertex nomination schemes on the basis of effectiveness (i.e.~precision) and also
computational practicality, as the scale of the number of vertices changes.
The colors red, blue, green, purple, and orange correspond respectively to the
canonical $\mathcal{L}^C$, canonical sampling  $\mathcal{L}^{CS}$, likelihood maximization $\mathcal{L}^{ML}$,
extended spectral partitioning $\mathcal{L}^{EP}$, and spectral partitioning $\mathcal{L}^P$ vertex nomination schemes.
The lines dim to reflect increased computational burden. The red line on top represents the
canonical vertex nomination scheme $\mathcal{L}^C$; it quickly dims out at a few tens of vertices,
since at this point $\mathcal{L}^C$ is no longer practical to compute. Otherwise, the red line
would have extended in a straight line across the figure, above all of the other lines, since
it is the optimal nomination scheme (in the sense of precision), and is thus the benchmark for comparison of all of the other
nomination schemes. Next, the dark/light/lighter blue regions correspond to the canonical sampling
vertex nomination scheme $\mathcal{L}^{CS}$; it isn't a single line, but rather layers of lines
for the different amounts of sampling that could be performed. As the number of vertices grows,
$\mathcal{L}^{CS}$ requires more sampling---i.e.~computational burden---to be more effective, hence the blue color
lightens upwards in the figure, as it approaches the red line---or where the red line would
have extended to. For graphs with few vertices, the dark blue line is just below the red line;
indeed, the canonical sampling scheme is just as effective as the canonical scheme, and without much computational burden.
Even with more vertices, with enough sampling we would have $\mathcal{L}^{CS}$ approaching
$\mathcal{L}^C$, but with an
ever increasing computational burden, hence the dimming of the blue towards the top of the figure. Next, the green line corresponds
to the likelihood maximization vertex nomination scheme $\mathcal{L}^{ML}$; the green color dims out
at a few thousand vertices, since at this point it is no longer practical to compute. Finally, the
purple and orange lines, respectively, correspond to the extended spectral partitioning $\mathcal{L}^{EP}$,
and spectral partitioning $\mathcal{L}^P$ vertex nomination schemes, the former being uniformly
more effective then the latter. When there are only a few vertices the spectral methods are
essentially useless, and these methods only become effective when there are a moderate number of
vertices. The extended spectral partitioning scheme is practical to compute until there are close
to a hundred thousand vertices, while the spectral partitioning scheme is practical to compute
even for many millions
of vertices.

The paper is laid out as follows.
In Section \ref{S:canon}, we describe the canonical vertex nomination scheme, and
prove its theoretical optimality in a slightly different model setting than
considered in \cite{fishkind2015vertex}.
In Section \ref{S:canonsamp}, we use Markov chain Monte Carlo methods to extend the
canonical vertex nomination scheme
$\mathcal{L}^C$ to the canonical sampling vertex nomination scheme
$\mathcal{L}^{CS}$.
In Section \ref{Sec:spec}, we describe the spectral partitioning nomination scheme.
In Section \ref{Sec:ssclust}, we extend the
spectral partitioning vertex nomination scheme $\mathcal{L}^P$ to
the extended spectral partitioning vertex nomination scheme $\mathcal{L}^{EP}$,
utilizing a more sophisticated clustering methodology than in $\mathcal{L}^P$,
without an inordinately large sacrifice in scalability.
In Section \ref{sec:data}, we demonstrate and compare the performance of
$\mathcal{L}^{EP}$ and $\mathcal{L}^{CS}$ on both simulated  and real data sets.\\

\section{Setting}
\label{Setting}

We develop our vertex nomination schemes in the setting of the stochastic block model,
a random graph model extensively used to model networks with underlying community structure.
See, for example, \cite{Holland1983,Wang1987,Airoldi2008}. The stochastic block model is a
very simple random graph model that provides a principled
approximation for more complicated network data (see, for example,
\cite{olhede_wolfe_histogram,wolfe_olhede_graphon,karrer11:_stoch}), with the
advantage that the  theory associated with the stochastic block model is quite tractable.

The stochastic block model random graph is defined as follows; let $K$ be a fixed positive integer.
\begin{defn}
A random graph ${\bf G}$ is an SBM$(K,\vec{n},b,\Lambda)$ graph if
\begin{itemize}
\item[i.] The vertex set $V$ is the disjoint union of $K$ sets
$V=V_1 \sqcup V_2 	\sqcup \cdots \sqcup V_K$ such that, for each $i=1,2,\ldots,K$,
it holds that $|V_i|=n_i$. (For each $i$, $V_i$ is called the $i$th {\bf block}.)
\item[ii.] The {\bf block membership function} $b:V \rightarrow \{1,2,\ldots,K\}$ is such that, for
all $v \in V$ and all $i=1,2,\ldots,K$, it holds that  $b(v)=i$ if and only if $v\in V_i$.
\item[iii.] The {\bf Bernoulli matrix} $\Lambda\in (0,1)^{K\times K}$ is such that, for each pair
of vertices $\{u,v\}\in\binom{V}{2}$, there is an edge between $u$ and $v$ (denoted $u \sim_{\bf G} v$)
with probability $\Lambda_{b(u),b(v)}$, and the collection of indicator random variables
$\{\grr_{u\sim_{\bf G} v}\}_{\{u,v\}\in\binom{V}{2}}$
is independent.
\end{itemize}
\end{defn}

In the setting of vertex nomination, we assume that $b$ is only partially observed.
Specifically, $V$ is partitioned into two disjoint sets, $S$
(the set of {\it seeds}) and $A$ (the set of {\it ambiguous vertices}), and we assume
that the values of $b$ are known only on $S$. We denote the restriction of
$b$ to $S$ as $\br :S \rightarrow \{1,2,\ldots,K\}$. For each $i=1,2,\ldots,K$,
we denote $A_i:=V_i \cap A$, \ $S_i:=V_i \cap S$, \ $m_i=|S_i|$, then we define $m:=\sum_{i=1}^Km_i$,
and $n:=\sum_{i=1}^Kn_i$. Of course, $|S|=m$ and $|A|=n-m$.

Given an SBM$(K,\vec{n},b,\Lambda)$ model where the parameters are unknown, these parameters can
be approximated in all of the usual ways
utilizing a graph $G$ realized from ${\bf G}\sim$SBM$(K,\vec{n},b,\Lambda)$. First,
$K$ can be consistently estimated by spectral methods
(such as in \cite{fishkind2013consistent,wang2017likelihood}).
Alternatively, since $\br$ is observed, we would be observing $K$
if we knew that $\br$ was a surjective function.
Given $K$, and assuming that the vertex memberships were realized via a multinomial distribution, then
$n_i$ can be estimated by $\frac{m_i}{m}n$, for each $i=1,2,\ldots,K$. Then,
for any $i,j \in \{1,2,\ldots,K\}$ such that $i\ne j$, we can estimate
$\Lambda_{i,j}$ by the number of edges in the bipartite subgraph induced by
$S_i,S_j$, divided by $m_im_j$; i.e.,

\begin{align}
\label{paraest1}
\widehat\Lambda_{i,j}=\frac{|\{ \{u,v\}\in E\text{ s.t. }u\in S_i,\,v\in S_j\}|}{m_im_j}.
\end{align}

For~$i=j$, we can estimate $\Lambda_{i,i}$
by the number of edges in the subgraph induced by $S_i$, divided by $\binom{m_i}{2}$; i.e.,

\begin{align}
\label{paraest2}
\widehat\Lambda_{i,i}=\frac{|\{ \{u,v\}\in E\text{ s.t. }u,v\in S_i\}|}{\binom{m_i}{2}}.
\end{align}

In simulations, when it useful or simplifying to do so, we assume that the
model parameters $K$, $\vec{n}$, $\Lambda$ are known.
Else, they are estimated as above.

Next, the most general inference task here would be,
given observed $G$ from ${\bf G}\sim$SBM$(K,\vec{n},b,\Lambda)$ and a
partially observed block membership function $\br$, to estimate the parameter $b$; that is,
to estimate the remaining unobserved block memberships. Indeed, there are a host of graph clustering algorithms that
could be used for this purpose; see, for example,
\cite{rohe2011spectral,qin2013dcsbm,STFP,bickel2013asymptotic,newman2006modularity,von2007tutorial} among others.
However, in the vertex nomination \cite{marchette2011vertex,coppersmith2012vertex,sun2012comparison,coppersmith2014vertex,fishkind2015vertex}
setting of this manuscript, the task  of interest is much more specialized. We assume that there is only one
block ``of interest"---without loss of generality
it is $V_1$---and we want to prioritize ambiguous vertices per the
possibility of being from $V_1$. Specifically, our task is, given an observed $G$ and
a partially observed block membership function $\br$, to order the
ambiguous vertices $A$ into a list such that there would be an abundance of vertices
from $V_1$ that appear as near to the top of the list as can be achieved. More formally:
\begin{defn} Given $S,\ A,$ and $ \br$,
a \textup{vertex nomination scheme} ${\mathcal L}$ is a function ${\mathcal L}:\mathcal{G}  \mapsto A!$ where
$\mathcal{G}$~is~the set of all graphs on vertex set $V=S \sqcup A$, and  $A!$ is the set of all orderings of the set $A$.
For any given $G \in \mathcal{G} $, denote the ordering $\mathcal{L}(G)$ of $A$
as $(\mathcal{L}_{G,1}, \mathcal{L}_{G,2}, \ldots \mathcal{L}_{G,n-m})$;
this ordering is also called the \textup{nomination list} associated with ${\mathcal L}$ and $G$. \label{def:nomlist}
\end{defn}

As in \cite{fishkind2015vertex}, it is helpful for analysis
to assume that for all graphs with symmetry (i.e., when a graph has a
nontrivial automorphism group), that all vertex nomination
schemes ${\mathcal L}$ assign
such graphs to an empty nomination list.
There isn't much loss of generality in this, since the number of graphs
with symmetry is very quickly negligible as the number of vertices increases \cite{ER1963,Polya1937}. We also require that
all vertex nomination schemes ${\mathcal L}$ have the following property:
For any asymmetric $G,H\in \mathcal{G}$ such that $G$ is isomorphic to $H$
via isomorphism $\gamma$ such  that $\gamma$ is the identity function on $S$, we require that
 $\gamma ({\mathcal L}_{G,i})
={\mathcal L}_{H,i} $ for all $i$. In  words,
${\mathcal L}$ should order the ambiguous vertices as if they are unlabeled.

The effectiveness of a vertex nomination
scheme ${\mathcal L}$ is quantified in the following manner.
Given a realization $G$ of ${\bf G} \sim $SBM$(K,\vec{n},b,\Lambda)$ and the partially observed block membership function $b$,
and for any integer $j=1,2,\ldots,n-m$,
define the {\it precision at depth~$j$} of the list ${\mathcal L}(G)$ to be

\begin{align*}
\frac{ | \{ i \mbox{ such that } 1 \leq i \leq j, \ b({\mathcal L}_{G,i})=1\}
| }{j} \ \ ;
\end{align*}

\noindent that is, the fraction of the first $j$ vertices
on the nomination list that are in the block of interest,~$V_1$.
The {\it average precision} of the list ${\mathcal L}(G)$ is defined to be the
average of the precisions at depths $j=1,2,\ldots,n_1-m_1$; that is, it is equal to
\begin{eqnarray}
\frac{1}{n_1-m_1} \sum_{j=1}^{n_1-m_1}
\frac{ | \{ i \mbox{ such that } 1 \leq i \leq j, \ b({\mathcal L}_{G,i})=1 \}
| }{j}.
\end{eqnarray}

Of course, average precision is defined for a particular instantiation of $G$, and
hence does not capture the behavior of $\mathcal{L}$ as $G$ varies in the SBM model.
To account for this, we define the {\it mean average precision}, the metric by which we will evaluate our vertex nomination schemes:
\begin{defn}
\label{defn:MAP}
Let ${\bf G}\sim$SBM$(K,\vec{n},b,\Lambda)$.
The {\it mean average precision} of a vertex nomination scheme
$\mathcal{L}$ is defined to be
\begin{align*}
MAP(\mathcal{L})&=\e  \left( \frac{1}{n_1-m_1} \sum_{j=1}^{n_1-m_1}
\frac{ | \{ i \textup{ such that } 1 \leq i \leq j, \ b({\mathcal L}_{{\bf G},i})=1 \}
| }{j} \right),
\end{align*}
where the expectation is taken over the underlying probability space, the sample space being ${\mathcal G}$.
\end{defn}

\noindent It is immediate that, for any given vertex nomination scheme $\mathcal{L}$,
the mean average precision satisfies $MAP(\mathcal{L})\in[0,1]$, with values closer to $1$
indicating a more successful nomination scheme; i.e., a higher concentration of
vertices from $V_1$ near the top of the nomination list.

In the literature, mean average precision is often defined as the integral of the precision over recall.
%
%
Herein, we focus on the definition of mean average precision
provided in Definition \ref{defn:MAP} because, in the vertex nomination
setting, recall is not as important as precision; the goal is explicitly
to have an abundance of vertices of interest at the top of the
list, and less explicitly about wanting all the vertices of interest to be high in the list.


\section{Extending the vertex nomination schemes}

In this section, we extend the canonical vertex nomination scheme
${\mathcal L}^{C}$ (described in Section \ref{S:canon}) to a ``sampling" version~${\mathcal L}^{CS}$
(defined in Section~\ref{S:canonsamp}), and we extend
the spectral partitioning vertex nomination scheme ${\mathcal L}^{P}$
(described in Section \ref{Sec:spec}) to ${\mathcal L}^{EP}$
(defined in Section \ref{Sec:ssclust}).


\subsection{The canonical vertex nomination scheme ${\mathcal L}^{C}$ }
\label{S:canon}

The canonical vertex nomination scheme ${\mathcal L}^C$, introduced in the paper \cite{fishkind2015vertex},
is defined to be the vertex nomination scheme which orders
the ambiguous vertices of  $A$ according to the order
of their conditional probability---conditioned on $G$---of being members
of the block of interest $V_1$. Indeed, it is intuitively clear
why this would be an excellent (in fact, optimal) nomination scheme.
However, since $b$ is a parameter,
this conditional probability is not yet meaningfully defined.
We therefore expand the probability space of the SBM
model given in Section \ref{Setting}, and construct a probability measure $\q$
for which the canonical vertex nomination scheme  ${\mathcal L}^C$ can be
meaningfully defined, with its requisite
conditional probabilities.
The probability measure $\q$~is~constructed~as~follows:

Define $\Phi$ to be the collection of functions $\varphi: V
\rightarrow \{ 1,2,\ldots, K \}$ such that $\varphi (v) = b (v)$ for
all $v \in S$, and such that $\Big | \{ v \in V : \varphi (v)=i  \} \Big |=n_i$ for all $i=1,2,\ldots,K$.
Also, recall that ${\mathcal G}$ is the set of all graphs on $V$.
The probability measure $\q$ has sample space ${\mathcal G}\times \Phi $,
and it is sampled from by first choosing $\varphi \in \Phi$ discrete-uniform
randomly and then, conditioned on $\varphi$, $G$ is chosen from the distribution
SBM$(K,\vec{n},\varphi,\Lambda)$. So, for all $G \in {\mathcal G}$, $\varphi \in \Phi$,

\begin{align}
\q (G,\varphi)&= \frac{1}{  \binom{ n-m}{n_1-m_1, n_2-m_2, \ldots, n_K-m_K}
}\prod_{i=1}^K\prod_{j=i}^K \left ( \Lambda_{i,j} \right )^{e_{i,j}^{G,\varphi}}
\left ( 1- \Lambda_{i,j}   \right )^{ c_{i,j}^{G,\varphi}} ,
\end{align}

\noindent where
$e_{i,j}^{G,\varphi}$ is defined  as the number of edges in $G$ such that $\varphi$ of one endpoint is $i$ and
$\varphi$ of the other endpoint is $j$, and we define $c_{i,j}^{G,\varphi}:=
n_in_j-e_{i,j}^{G,\varphi}$ if $i\neq j$, and $c_{i,i}^{G,\varphi}:=\binom{n_i}{2}-e_{i,i}^{G,\varphi}$.
This probability measure, with uniform marginal distribution on $\Phi$,
reflects our situation  where we have
no prior knowledge of specific block membership for the ambiguous vertices (beyond
block sizes). Note that $\q$ is an intermediate measure used to show that
$\mathcal{L}^C$ is optimal as stated in Theorem \ref{thm:optimal}.

The first step in the canonical nomination scheme is to update this uniform distribution
on $\Phi$ to reflect what is learned from the realization of the graph.
Indeed, conditioning on any $G \in {\mathcal G}$, the conditional sample space of $\q$ collapses to become
$\Phi$ and, for any $\varphi \in \Phi$, we have by Bayes Rule that
\begin{eqnarray} \label{cond}
\q (\varphi|G) = \frac{ \q (G,\varphi)}{\sum_{\psi\in \Phi} \q (G,\psi)}=\frac{
\prod_{i=1}^K\prod_{j=i}^K \left ( \Lambda_{i,j} \right )^{e_{i,j}^{G,\varphi}}
\left ( 1- \Lambda_{i,j}   \right )^{c_{i,j}^{G,\varphi}}     }{
\sum_{\psi \in \Phi }  \prod_{i=1}^K\prod_{j=i}^K \left ( \Lambda_{i,j} \right )^{e_{i,j}^{G,\psi}}
\left ( 1- \Lambda_{i,j}   \right )^{c_{i,j}^{G,\psi }}        }.
\end{eqnarray}
In all that follows in this subsection,
let ${\bf G}, \phi $ respectively denote the random graph and the random function, together distributed as $\q$; in particular, the random
${\bf G}$ is ${\mathcal G}$-valued, and the random $\phi$ is $\Phi$-valued.
For each $v \in A$, the event $\phi(v)=1$ is the event $\{ \varphi \in \Phi : \varphi (v)=1 \}$ and, by~Bayes'~Rule,

\begin{eqnarray}\label{eq:Qforcanon}
\q ( \ \phi (v) = 1 \ \big | \ G \ ) = \frac{
\sum_{\varphi \in \Phi : \varphi (v)=1} \prod_{i=1}^K\prod_{j=i}^K \left ( \Lambda_{i,j} \right )^{e_{i,j}^{G,\varphi}}
\left ( 1- \Lambda_{i,j}   \right )^{c_{i,j}^{G,\varphi}}     }{
\sum_{\varphi \in \Phi }  \prod_{i=1}^K\prod_{j=i}^K \left ( \Lambda_{i,j} \right )^{e_{i,j}^{G,\varphi}}
\left ( 1- \Lambda_{i,j}   \right )^{c_{i,j}^{G,\varphi}}        }.
\end{eqnarray}

The canonical vertex nomination scheme ${\mathcal L}^C$ is then defined as ordering the vertices
in $A$ by decreasing value of $\q (\phi (v) = 1 |G)$ (with ties broken arbitrarily);

\begin{align}
{\mathcal L}^C_{G,1}&\in\text{argmax}_{v\in A} \q (\phi (v) = 1|G)\notag;\\
{\mathcal L}^C_{G,2}&\in\text{argmax}_{v\in A\setminus{\mathcal L}^C_{G,1}} \q (\phi (v) = 1 |G)\notag;\\
&\vdots\notag\\
{\mathcal L}^C_{G,n-m}&\in\text{argmax}_{v\in A\setminus\left(\cup_{j=1}^{n-m-1}{\mathcal L}^C_{G,j}\right)}\q (\phi (v) = 1 |G). \label{eq:canonorder}
\end{align}

In \cite{fishkind2015vertex} it is proved that the canonical vertex nomination scheme is an
optimal vertex nomination scheme, in the sense of Theorem \ref{canonicaloptimal}.
We include the proof of Theorem \ref{canonicaloptimal} to reflect changes in our setting  from the setting in \cite{fishkind2015vertex}.
Recall from the paragraph after Definition \ref{def:nomlist}
that we assume that all vertex nomination schemes assign graphs with symmetry to an empty nomination list.
There isn't much impact in this, since the number of graphs with symmetry is quickly negligible as the number of vertices increases \cite{ER1963,Polya1937}.
Then, for any asymmetric $G,H\in \mathcal{G}$ such that $G$ is isomorphic to $H$ via isomorphism $\gamma$ such  that $\gamma$ is the identity function on $S$,
we also required that $\gamma ({\mathcal L}_{G,i})
={\mathcal L}_{H,i} $ for all $i$; in  words,
${\mathcal L}$ should order the ambiguous vertices as if they are unlabeled. Clearly ${\mathcal L}^C$ satisfies this.

\begin{theorem}
\label{thm:optimal} For any stochastic block model SBM($K,\vec{n},b,\Lambda)$ and vertex
nomination scheme ${\mathcal L}$, it holds that $MAP({\mathcal L}^C) \geq MAP({\mathcal L})$.
\label{canonicaloptimal}
\end{theorem}

\noindent {\bf Proof of Theorem \ref{canonicaloptimal}:}
For each $i=1,2,\ldots,n_1-m_1$, define
$\alpha_i:=\frac{1}{n_1-m_1} \sum_{j=i}^{n_1-m_1}\frac{1}{j}$ and then,
for each of $i=n_1-m_1+1, \ n_1-m_1+2, \ \ldots,\ n-m$, define
$\alpha_i:=0$. Note that the sequence of $\alpha_i$'s is nonnegative and nonincreasing.
Thus, for any other nonnegative and nonincreasing sequence of real numbers
$a_1,a_2,\ldots,a_{n-m}$ and any rearrangement
$a'_1,a'_2,\ldots,a'_{n-m}$ of the sequence $a_1,a_2,\ldots,a_{n-m}$,
we have by the Rearrangement Inequality \cite{rearrange} that
\begin{eqnarray}   \label{first}
\sum_{i=1}^{n-m} \alpha_ia'_i \ \leq \ \sum_{i=1}^{n-m}\alpha_i a_i.
\end{eqnarray}

Next, consider any $\varphi,\varphi' \in \Phi$, and suppose that a
function $\gamma :V \rightarrow V $ is bijective, that $\gamma$ is the identity function on $S$, and that $\gamma$ satisfies $\forall v \in A$, $\varphi(v)=
\varphi'(\gamma(v))$. For any $G \in {\mathcal G}$, let $\gamma (G)$
denote the graph in ${\mathcal G}$ isomorphic to $G$ via the
isomorphism $\gamma$; it is clear that (under our assumptions,
in particular suppose $G$ is asymmetric) $\gamma ({\mathcal L}^C_{G,i})
={\mathcal L}^C_{\gamma(G),i} $ for all $i$, since the canonical
vertex nomination scheme orders the vertices as if they are unlabeled.
Thus, since $\gamma: {\mathcal G} \rightarrow {\mathcal G}$
is clearly bijective,
we have for all $i$ that

\begin{eqnarray*}
&  & \mathbb{Q} \left (
  \phi({\mathcal L}^C_{{\bf G},i})=1
 \mathlarger{\mathlarger{\mathlarger{|}}}  \ \phi = \varphi \  \right )
 =
 \sum_{v \in A \ : \ \varphi(v)=1}
 \mathbb{Q} \left (
  v={\mathcal L}^C_{{\bf G},i}
 \mathlarger{\mathlarger{\mathlarger{|}}}  \ \phi = \varphi \  \right )\\
 &=&
 \frac{1}{  \binom{ n-m}{n_1-m_1, n_2-m_2, \ldots, n_K-m_K}} \
 \sum_{v \in A \ : \ \varphi(v)=1} \ \
 \sum_{G \in {\mathcal G}\ : \ v= {\mathcal L}^C_{G,i}} \mathbb{Q}(G,\varphi) \\
 &=&
 \frac{1}{  \binom{ n-m}{n_1-m_1, n_2-m_2, \ldots, n_K-m_K}} \
 \sum_{v \in A \ : \ \varphi'(v)=1} \ \
 \sum_{G \in {\mathcal G}\ : \ v= {\mathcal L}^C_{\gamma(G),i}} \mathbb{Q}(\gamma(G),\varphi') \\
 &=& \sum_{v \in A \ : \ \varphi'(v)=1}
 \mathbb{Q} \left (
  v={\mathcal L}^C_{{\bf G},i}
 \mathlarger{\mathlarger{\mathlarger{|}}}  \ \phi = \varphi' \  \right )
 =
 \mathbb{Q} \left (
  \phi({\mathcal L}^C_{{\bf G},i})=1
 \mathlarger{\mathlarger{\mathlarger{|}}}  \ \phi = \varphi' \  \right );
\end{eqnarray*}
since $\varphi$ and $\varphi'$ were arbitrary, the preceding is thus equal
to (unconditioned)
$\mathbb{Q} \left (
  \phi({\mathcal L}^C_{{\bf G},i})=1 \right )$.
Hence, for all $i$, we have that
\begin{eqnarray} \label{second}
\mathbb{Q} \left (
  b({\mathcal L}^C_{{\bf G},i})=1
 \mathlarger{\mathlarger{\mathlarger{|}}}  \ \phi = b \      \right )
 =
 \mathbb{Q} \left (
  \phi({\mathcal L}^C_{{\bf G},i})=1     \right ) .
\end{eqnarray}
By the same reasoning,
 the vertex nomination scheme ${\mathcal L}$ also satisfies Equation
 (\ref{second}).

Now, to the main line of reasoning in the proof:
\begin{eqnarray*} \label{longish}
MAP(\mathcal{L}^C)&=& \e  \left( \frac{1}{n_1-m_1} \sum_{j=1}^{n_1-m_1}
\frac{ | \{ i \textup{ such that } 1 \leq i \leq j, \ b({\mathcal L}^C_{{\bf G},i})=1 \}
| }{j} \ \mathlarger{\mathlarger{\mathlarger{|}}}  \ \phi = b \     \right)\\
 & = &  \e \left ( \sum_{i=1}^{n-m}  \alpha_i \cdot
 \grr \left [ (b({\mathcal L}^C_{{\bf G},i})=1 \right ]
 \mathlarger{\mathlarger{\mathlarger{|}}}  \ \phi = b \      \right )     \\
 & = &  \sum_{i=1}^{n-m}  \alpha_i \cdot \mathbb{Q} \left (
  b({\mathcal L}^C_{{\bf G},i})=1
 \mathlarger{\mathlarger{\mathlarger{|}}}  \ \phi = b \      \right )     \\
 & = &  \sum_{i=1}^{n-m}  \alpha_i \cdot \mathbb{Q} \left (
  \phi({\mathcal L}^C_{{\bf G},i})=1     \right )
  \ \ \ \ \ \ \ \ \ \ \ \ \ \ \ \ \
  \ \ \ \ \ \ \ \ \ \ \ \ \ \ \ \ \ \ \ \mbox{by Equation (\ref{second}})\\
  & = &  \sum_{i=1}^{n-m}  \alpha_i \left ( \sum_{G \in {\mathcal G}}
  \mathbb{Q}(G) \cdot \q  \left (
  \phi({\mathcal L}^C_{G,i})=1  \
  \mathlarger{\mathlarger{|}} \ G \ \right ) \right ).
   \end{eqnarray*}
   From this, we have
    \begin{eqnarray*}
    \label{longish2}
  MAP(\mathcal{L}^C)
  & = & \sum_{G \in {\mathcal G}}
  \mathbb{Q}(G)
   \sum_{i=1}^{n-m}  \alpha_i   \cdot \q  \left (
  \phi({\mathcal L}^C_{ G,i})=1  \
  \mathlarger{\mathlarger{|}} \ G \ \right )     \\
 & \geq & \sum_{G \in {\mathcal G}}
  \mathbb{Q}(G)
   \sum_{i=1}^{n-m}  \alpha_i   \cdot \q  \left (
  \phi({\mathcal L}_{G,i})=1  \
  \mathlarger{\mathlarger{|}} \ G \ \right )
  \ \ \ \ \ \ \ \  \mbox{by definition of ${\mathcal L}^C$, Equation (\ref{first}})
      \\
   & = &  \sum_{i=1}^{n-m}  \alpha_i \cdot \mathbb{Q} \left (
  \phi({\mathcal L}_{{\bf G},i})=1     \right )     \\
  & = &  \sum_{i=1}^{n-m}  \alpha_i \cdot \mathbb{Q} \left (
  b({\mathcal L}_{{\bf G},i})=1
 \mathlarger{\mathlarger{\mathlarger{|}}}  \ \phi = b \      \right ) \\  &=& \e  \left( \frac{1}{n_1-m_1} \sum_{j=1}^{n_1-m_1}
\frac{ | \{ i \textup{ such that } 1 \leq i \leq j, \ b({\mathcal L}_{{\bf G},i})=1 \}
| }{j} \ \mathlarger{\mathlarger{\mathlarger{|}}}  \ \phi = b \     \right)\\
& =& MAP(\mathcal{L}),
 \end{eqnarray*}
 which completes the proof of Theorem \ref{canonicaloptimal}. \qed


\subsection{The canonical sampling vertex nomination scheme  ${\mathcal L}^{CS}$   }
\label{S:canonsamp}

The formula in Equation \ref{eq:Qforcanon}  can be directly
used to compute $\q (\phi (v) = 1 |G)$ for all $v \in A$,
to obtain the ordering that defines the canonical vertex nomination scheme ${\mathcal L}^C$, but due to
 the burgeoning number of summands in the numerator and in the
 denominator of Equation \ref{eq:Qforcanon}, this direct approach is
 computationally intractable, feasible only when the number of vertices
is on the order of a few tens.
We next introduce an extension of the canonical vertex nomination
scheme called the {\it canonical sampling vertex nomination scheme}
${\mathcal L}^{CS}$.
The purpose of the canonical sampling vertex nomination scheme is to provide a
computationally tractable estimate
$\widehat{\q} (\phi (v) = 1 |G)$ of
$\q (\phi (v) = 1 |G)$, for all $v \in A$.
The nomination
list for ${\mathcal L}^{CS}$ consists of the vertices $v \in A$ ordered
by nonincreasing values of $\widehat{\q} (\phi (v) = 1 |G)$, exactly as
the nomination
list for ${\mathcal L}^{C}$ consists of the vertices $v \in A$ ordered
by nonincreasing values of $\q (\phi (v) = 1 |G)$.

Given the realized graph instance $G \in {\mathcal G}$ of the random graph ${\bf G}$,
we obtain the approximation
$\widehat{\q} (\phi (v) = 1 |G)$ of $\q (\phi (v) = 1 |G)$ for all $v \in A$
by sampling from the conditioned-on-$G$ probability space
$\q ( \cdot |G)$ on $\Phi$, then, for each $v \in A$,
$\widehat{\q} (\phi (v) = 1 |G)$ is defined as the fraction of the sampled functions
($\Phi$ is a set of functions) that map $v$ to the integer $1$.
The formula for the conditional probability distribution $\q ( \cdot |G)$
is given in Equation~\ref{cond}; unfortunately,
straightforward sampling from this distribution is hampered
by the intractability of directly computing the denominator of Equation~\ref{cond}.
Fortunately, sampling in this setting can be achieved via Metropolis-Hastings
Markov chain Monte Carlo.
For relevant background on Markov chain Monte Carlo, see, for example,
\cite[Chapter 11]{gelman2014bayesian} or \cite[Chapter 11]{aldous2002reversible}.

The base chain that we employ in our Markov chain Monte Carlo approach is the
well-studied Bernoulli-Laplace diffusion model \cite{feller2008introduction}.
The state space for the Markov chain is the set $\Phi$, and the one-step transition
probabilities, denoted  P$(\cdot, \cdot)$, for this chain are defined,
for all $\varphi,\varphi' \in \Phi$, as

\begin{align*}
\tP(\varphi,\varphi')=\frac{
\grr \{d(\varphi,\varphi')=2\}}{  \binom{n-m}{2} -\sum_{i=1}^K\binom{n_i-m_i}{2}   } ,
\end{align*}

\noindent
where $d(\varphi,\varphi'):= |\{v\text{ such that }\varphi(v)\neq\varphi'(v)\} |$.
In other words, if at state $\varphi$, a move transpires as follows.
A pair of vertices $\{ u,v \} \in \binom{A}{2}$ is chosen
discrete-uniformly at random, conditional on the fact that $\varphi(u)\ne \varphi (v)$,
and then the move is to state $\varphi'$, which is defined
as agreeing with $\varphi$, except that $\varphi'(u)$ and $\varphi'(v)$ are defined
respectively as $\varphi(v)$ and $\varphi(u)$.
We will see shortly that the simplicity of this base chain greatly simplifies the computations
needed to employ Metropolis-Hastings with target distribution $\q (\cdot|G)$.

The Metropolis-Hastings chain has state space $\Phi$, and one-step transition
probabilities, $\widehat\tP (\cdot,\cdot)$ defined
for all $\varphi,\varphi' \in \Phi$ as

\begin{align*}
\widehat\tP(\varphi,\varphi')&=\frac{
\grr \{d(\varphi,\varphi')=2\}}{  \binom{n-m}{2} -\sum_{i=1}^K\binom{n_i-m_i}{2}   }\min \left \{ \ \  1, \ \
\frac{\prod_{i=1}^K\prod_{j=i}^K \left ( \Lambda_{i,j} \right )^{e_{i,j}^{G,\varphi'}}
\left ( 1- \Lambda_{i,j}   \right )^{c_{i,j}^{G,\varphi'}} }{\prod_{i=1}^K\prod_{j=i}^K \left ( \Lambda_{i,j} \right )^{e_{i,j}^{G,\varphi}}
\left ( 1- \Lambda_{i,j}   \right )^{c_{i,j}^{G,\varphi}} } \ \
\right \}\text{ if }\varphi \neq \varphi';\\
\widehat\tP(\varphi,\varphi)&=1-\sum_{\varphi'' \in \Phi : \varphi'' \neq \varphi}\widehat\tP(\varphi,\varphi'').
\end{align*}

\noindent
In other words, if at state $\varphi$, a candidate state $\varphi'$ is proposed according to
$\tP(\varphi,\cdot)$ and is independently accepted as the next state of the Markov chain with
probability $\min \left \{ 1, \frac{\q (\varphi' |G)}{\q (\varphi |G)}
\right \}$.
It is immediate that the stationary distribution for $\widehat\tP$ is
$\q (\cdot|G)$ and that the chain is reversible with respect to $\q (\cdot|G)$;
that is, for any $\varphi,\varphi' \in \Phi$, \ \
$\q (\varphi |G ) \cdot \widehat\tP (\varphi, \varphi')  =
\q (\varphi' |G ) \cdot \widehat\tP (\varphi' , \varphi)$.

Note that
the simplicity of the underlying base chain greatly aids in the speedy
computation of $\frac{\q (\varphi' |G)}{\q (\varphi |G)}$
during the computation of transition probabilities $\widehat\tP$. Indeed,
since $\varphi$ and $\varphi'$ for which we might
want to compute $\widehat\tP(\varphi,\varphi')$
are such that $d(\varphi,\varphi')=2$, we would have that $\varphi$ and $\varphi'$ differ only
on two vertices, call them $u$,$v$, and  say that $i$ and $j$ are
such that $\varphi(u)=i$ and $\varphi(v)=j$.
Then
\begin{eqnarray} \label{update}
\frac{\q (\varphi' |G)}{\q (\varphi |G)}=\prod_{w \in V: w \ne u, w \ne v} \left (
\frac{\Lambda_{\varphi(w),j}(1-\Lambda_{\varphi(w),i})}{\Lambda_{\varphi(w),i}(1-\Lambda_{\varphi(w),j})} \right ) ^{\tiny
 \left \{ \begin{array}{rl} 1 & \mbox{ if }  w \sim_G u, w \not \sim_G v  \\  -1 & \mbox{ if }  w \not \sim_G u, w \sim_G v \\ 0 & \mbox{ else }
 \end{array} \right \} } .
\end{eqnarray}
This reduces the number of operations to compute $\frac{\q (\varphi' |G)}{\q (\varphi |G)}$ from
$O(n^2)$ down to $O(n)$. As an implementation note,
in practice we would utilize a logarithm to convert Equation \ref{update} from a multiplicative
expression into an additive expression, which will greatly reduce roundoff error
that can arise when working with numbers that are orders of magnitude different from each other.

Now, the canonical sampling vertex nomination scheme ${\mathcal L}^{CS}$ is defined in the
exact same manner as ${\mathcal L}^C$, except that, for all $v \in A$,
the value $\widehat{\q} (\phi (v) = 1 |G)$ is approximated as follows.  Denoting the
Metropolis-Hastings Markov chain by $(X_t)_{t=0}^\infty$, we set $X_0\sim$Uniform$(\Phi$).
After evolving the chain past a ``burn-in" period, $T$, we approximate $\q (\phi (v) = 1 |G)$ via

$$\widehat \q (\phi (v) = 1 |G)=\frac{|\{s\text{ such that }T<s \leq T+t,\text{ and }X_s(v)=1\}|}{t} , $$

\noindent for a predetermined number of Metropolis-Hastings steps $t.$
For fixed $T$, we then have as an immediate consequence of the Ergodic Theorem (see, for example,
\cite[Chp.~2, Thm.~1]{aldous2002reversible})
that $\lim_{t\rightarrow\infty}\widehat \q (\phi (v) = 1 |G)=\q (\phi (v) = 1 |G)$ for each $v\in A$ (indeed, our Metropolis-Hastings chain is aperiodic, recurrent and finite state).

In this paper,
we do not address how to choose a suitable burn-in $T$ for a given
implementation of $\mathcal{L}^{CS}$, instead focusing on a feasible burn-in given limited computational resources.
Practically, there are a bevy of methods for approximating $T$, see for example those in \cite{gelman2014bayesian,gilks1995markov}.
Regarding mixing time, there is an unfortunate dearth of
rigorous mixing time computations for general, non-unimodal
Metropolis-Hastings algorithms (see the discussion in  \cite{Diaconis98,Johndrow17}), and such analysis is beyond
the scope of this paper. Our choice of the Bernoulli-Laplace base chain is
for its fast and efficient implementation of the sampling procedure,
although we have no guarantee or expectation of optimal mixing time.

\subsection{The spectral partitioning vertex nomination scheme}
\label{Sec:spec}

We now review the spectral partitioning vertex
nomination scheme $\mathcal{L}^P$ from \cite{fishkind2015vertex};
afterwards, in Section \ref{Sec:ssclust}, $\mathcal{L}^P$ will be extended to the
 vertex nomination scheme $\mathcal{L}^{EP}$.

As in Section \ref{Setting}, we assume here that the graph
$G$ is realized from an  SBM$(K,\vec{n},b,\Lambda)$ distribution,
where $K$ is known.
Furthermore, we assume that the values of the block membership function $b$ are
known only on the set of seeds $S$, and are not known on the set of ambiguous vertices $A=V \backslash S$.
In contrast to Section~\ref{S:canon}, here we
do not need to assume that $\vec{n}$ and $\Lambda$ are
explicitly known or estimated, except that $d:=\textup{rank }\Lambda$
is known, or an upper bound for $d$ is known.
As before, say that $V_1$ is the ``block~of~interest."

The spectral partitioning vertex nomination scheme $\mathcal{L}^P$
is computed in three stages; first is the adjacency spectral embedding of $G$, then
clustering of the embedded points, and then ranking the ambiguous vertices into the nomination list.
(The first two of these stages are collectively called
{\it adjacency spectral clustering}; for a good reference, see~\cite{von2007tutorial}.)
We begin by describing the first stage, adjacency spectral embedding:
\begin{defn}
Let graph $G$ have adjacency matrix ${\mathcal A}$, and suppose $({\mathcal A}^\top {\mathcal A})^{1/2}$ has
eigendecompostion
$$({\mathcal A}^\top {\mathcal A})^{1/2}=\big[U|\tilde U\big]\big[D\oplus \tilde D\big]\big[U|\tilde U\big]^\top;$$ i.e., $U \in \mathbb{R}^{n \times d}$,
$\big[U|\tilde U\big]\in\mathbb{R}^{n\times n}$ is orthogonal,
$[D\oplus \tilde D\big]\in\mathbb{R}^{n\times n}$ is diagonal,
and the diagonal of $D\in R^{d \times d}$ is composed of the
$d$ greatest eigenvalues of $({\mathcal A}^\top {\mathcal A})^{1/2}$
in nonincreasing order.
The $d$-dimensional \emph{adjacency spectral embedding} of $G$ is then given by $\widehat X=UD^{1/2}.$
In particular, for each $v\in V$, the row of $\widehat X$ corresponding to $v$,
denoted $\widehat X_v$, is the embedding of $v$ into $\mathbb{R}^{d}$.
\end{defn}

After the adjacency spectral embedding, the second stage
is to cluster the embedded vertices---i.e.~the associated
points in $\mathbb{R}^{d}$---using the
$k$-means clustering algorithm \cite{macqueen1967some}. The clusters so obtained
are estimates of the different blocks, and the cluster containing the most
vertices from $S_1:=S \cap V_1$ is an estimate of the block of interest $V_1$;
let $c$ denote the centroid of this cluster.
(Note that this clustering step, as described here for $\mathcal{L}^P$, is fully unsupervised,
not taking advantage of the observed memberships of the vertices in $S$.
In Section \ref{Sec:ssclust}, incorporating these labels
into a semi-supervised clustering step is a natural way to extend $\mathcal{L}^P$
and improve performance.)

The third stage is ranking the ambiguous vertices into the nomination list;
the vertices are nominated based on their Euclidean distance from $c$, the centroid
of the cluster which is the estimate for the block of interest. Specifically, define:

\begin{align}
\label{eq:schemespectral}
{\mathcal L}^P_{G,1}&\in\text{argin}_{v\in A}\|v-c\|_2;\notag\\
{\mathcal L}^P_{G,2}&\in\text{argmin}_{v\in A\setminus{\mathcal L}^P_{G,1}}\|v-c\|_2\notag;\\
&\vdots\notag\\
{\mathcal L}^P_{G,n-m}&\in\text{argmin}_{v\in A\setminus\left(\cup_{j=1}^{n-m-1}{\mathcal L}^P_{G,j}\right)}\|v-c\|_2.
\end{align}

For definiteness, any ties in the above procedure should be broken by choosing
uniform-randomly from the choices.
This concludes the definition of the spectral partitioning vertex nomination scheme $\mathcal{L}^P.$

Under mild assumptions, it is proven in \cite{perfect} that, in the limit,
adjacency spectral partitioning almost surely perfectly clusters the vertices of $G$
into the true blocks. This fact was leveraged in \cite{fishkind2015vertex}
to prove that if $m_1>0$ and there exists a $\gamma>0$ such that  for all $i=1,2,\ldots,K$,\ \
 $n_i\geq \gamma \cdot n^{3/4+\gamma}$, then $\lim_{n\rightarrow\infty}$MAP$(\mathcal{L}^P)=1.$

If $d$ is unknown, singular value thresholding \cite{chatterjee2014matrix} can be used
to estimate $d$ from a partial SCREE plot \cite{zhu2006automatic}.
We note that the results of \cite{fishkind2013consistent} suggest that there will
be little performance lost if $d$ is moderately overestimated.
Additionally, if $K$ is unknown then it can be estimated by optimizing the silhouette
width of the resulting clustering \cite{kaufman2009finding}.
A key advantage of the spectral nomination scheme is that, unlike $\mathcal{L}^C$,
$\Lambda$ and $\vec n$ need not be estimated before applying the scheme.

\subsection{The extended spectral partitioning vertex nomination scheme}
\label{Sec:ssclust}

In this section, we extend the spectral partitioning vertex
nomination scheme $\mathcal{L}^P$ (described
in the previous section) to the extended spectral partitioning vertex
nomination scheme $\mathcal{L}^{EP}$.
Just like in computing $\mathcal{L}^P$, computing the extended
spectral partitioning vertex nomination
scheme $\mathcal{L}^{EP}$ starts with adjacency
spectral embedding. Whereas the next stage of $\mathcal{L}^P$ is
unsupervised clustering using the k-means algorithm, $\mathcal{L}^{EP}$
will instead utilize a semi-supervised clustering procedure which we describe below.

There are numerous ways to incorporate the known block memberships for $S$
into the clustering step of adjacency spectral clustering
(see, for example, \cite{wagstaff2001constrained,yoder2014model}).
The results of \cite{athreya13} suggest that, for each vertex $v$ of ${\bf G}$,
the distribution of $v$'s embedding $\widehat X_v\in \mathbb{R}^{d} $ is approximately normal, with
parameters that depend only on which block $v$ is a member of,
and this normal approximation
gets closer to exact as $n$ grows.
We thus model ${\bf G}$'s embedded vertices
as independent draws from a $K$-component Gaussian mixture~model
(except for vertices of $S$, where the Gaussian component is specified);
~i.e.,~there exists a fixed nonnegative vector $\pi:=(\pi_1,\pi_2,
\ldots,\pi_K)\in \mathbb{R}^{K}$ satisfying $\sum_{k=1}^K\pi_k=1$, and for each $k=1,2,\ldots,K$,
there exists $\mu^{(k)} \in \mathbb{R}^{d}$ and $\Sigma^{(k)}
\in \mathbb{R}^{d \times d}$ such that, independently for each vertex $v \in A$, the
block of $v$ is $1,2,\ldots,K$ with respective probabilities
$\pi_1,\pi_2,\ldots,\pi_K$, and then, conditioning on model block membership---say
the block of $v$ is $k$---the distribution of $\widehat X_v $ is
Normal$(\mu^{(k)},\Sigma^{(k)})$, denote this density $f_{\mu^{(k)},\Sigma^{(k)}}$. If
$\boldsymbol{\mu}$ denotes the sequence of mean vectors $(\mu^{(1)},\mu^{(2)},\ldots,\mu^{(K)})$,
$\bsigma$ denotes the sequence of covariance matrices $(\Sigma^{(1)},\Sigma^{(2)},\ldots,\Sigma^{(K)})$,
and (random) $\varphi:V \rightarrow \{1,2,\ldots,K\}$
denotes the Gaussian mixture model block membership function---i.e., for each $v \in V$
and $k \in \{ 1,2,\ldots,K  \}$, it holds that $\varphi(v)=k$
precisely when the Gaussian mixture model places $v$ in block $k$---then
 the complete data log-likelihood function can be written as
\begin{eqnarray} \label{loglik}
\ell(\pi,\bmu,\bsigma)_{\widehat X,\varphi} =
\sum_{k =1}^K \sum_{v \in S_k}
 \log\left(
 f_{\mu^{(k)},\Sigma^{(k)}}(\widehat X_v)\right) +
\sum_{k =1}^K \sum_{v \in A}    \grr_{  \varphi (v)= k }
 \log\left(\pi_k f_{\mu^{(k)},\Sigma^{(k)}}(\widehat X_v)\right),
\end{eqnarray}
which meaningfully incorporates the seeding information contained in $S$.

If $\vec{n}$ is known (indeed, it was assumed to be known
in the formulation of $\mathcal{L}^{C}$, but was
not assumed to be known in the formulation of $\mathcal{L}^{P}$)
then, for each $k=1,2,\ldots,K$, we would substitute $\frac{n_k}{n}$
in place of $\pi_k$.

With this model is place, it is natural to cluster the rows of $\widehat X$ using a (semi-supervised)
Gaussian mixture model (GMM) clustering algorithm rather than (unsupervised) $k$-means employed by~$\mathcal{L}^P$.
We now return to the description of the extended spectral partitioning vertex
nomination scheme $\mathcal{L}^{EP}$ after the first stage---adjacency
spectral embedding---has been performed. The next stage---clustering---can
be cast as the problem of uncovering the latent $\grr_{  \varphi (v)= k }$'s as
are present in the log-likelihood in Equation~\ref{loglik}.
We employ a semi-supervised modification of the model-based \texttt{Mclust}
Gaussian mixture model methodology of \cite{fraley2002model,fraley2006mclust};
we call this modification \texttt{ssMclust}; note that \texttt{ssMclust} first
appeared in \cite{yoder2014model}, and we include a brief outline of its implementation
below for the sake of completeness.

As in \cite{fraley2002model}, \texttt{ssMclust} uses the expectation-maximization (EM) algorithm to
approximately find the maximum likelihood estimates of Equation \ref{loglik},
denote them by $\hat{\pi},\hat{\bmu},\hat{\bsigma}$.
For each $v \in A$, the cluster of $\widehat X_v$---which is an estimate for the block of  $v$---is then set to be
\begin{eqnarray*}
\widehat \varphi (v): =
\text{argmax}_{k \in \{ 1,2,\ldots, K \}}\hat{\pi}_k f_{\hat{\mu}^{(k)},\widehat{\Sigma}^{(k)}}(\widehat X_v).
\end{eqnarray*}
Details of the implementation of the semi-supervised EM algorithm can be found
in \cite{mclachlan2004finite,yoderdiss,yoder2014model}, and are omitted here for brevity.
We note here that we initialize the class assignments in the EM algorithm by
first running the semi-supervised $k$-means++ algorithm of \cite{yoder2015model} on $\widehat X$.
This initialization, in practice, has the effect of greatly reducing the running
time of the EM step in \texttt{ssMclust}; see \cite{yoderdiss}.
  \begin{table}[t!]

\centering
\tiny{\begin{tabular}{l l l l l l}
\hline\hline
Name  & Applicable To & $\Sigma^{(k)}$ & Volume & Shape & Orientation \\
\hline
E &  $\mathbb{R}$ & $\lambda$ &  Equal & NA & NA \\
V &  $\mathbb{R}$ & $\lambda_k$  &  Varying & NA & NA \\
X &  $\mathbb{R}, K=1$ & $\lambda$  & NA & NA & NA\\
EII &  $\mathbb{R}^d$ & $\lambda I$ & Equal & Equal, spherical & Coordinate axes \\
VII &  $\mathbb{R}^d$  & $\lambda_k I$ & Varying & Equal, spherical & Coordinate axes\\
EEI &  $\mathbb{R}^d$  & $\lambda D$ & Equal & Equal, ellipsoidal & Coordinate axes \\
VEI &  $\mathbb{R}^d$  & $\lambda_k D$ & Varying & Equal, ellipsoidal & Coordinate axes \\
EVI &  $\mathbb{R}^d$  & $\lambda D_k$ & Equal & Varying, ellipsoidal & Coordinate axes\\
VVI &  $\mathbb{R}^d$  & $\lambda_k D_k$ & Varying & Varying, ellipsoidal & Coordinate axes \\
EEE &  $\mathbb{R}^d$  & $\lambda U D U^T$ & Equal & Equal, ellipsoidal & Equal\\
EVE &  $\mathbb{R}^d$  & $\lambda U D_k U^T$  & Equal & Varying, ellipsoidal & Equal\\
VEE &  $\mathbb{R}^d$  & $\lambda_k U D U^T$  & Varying & Equal, ellipsoidal & Equal\\
VVE &  $\mathbb{R}^d$  & $\lambda_k U D_k U^T$  & Varying & Varying, ellipsoidal & Equal\\
EEV &  $\mathbb{R}^d$  & $\lambda U_k D U_k^T$  & Equal & Equal, ellipsoidal & Varying\\
VEV &  $\mathbb{R}^d$  & $\lambda_k U_k D U_k^T$  & Varying & Equal, ellipsoidal & Varying\\
EVV &  $\mathbb{R}^d$  & $\lambda U_k D_k U_k^T$  & Equal & Varying, ellipsoidal & Varying\\
VVV &  $\mathbb{R}^d$  & $\lambda_k U_k D_k U_k^T$  & Varying & Varying, ellipsoidal & Varying\\
XII &  $\mathbb{R}^d, K=1$ & $\lambda I$  & NA & Spherical & Coordinate Axes\\
XXI &  $\mathbb{R}^d, K=1$  & $\lambda D$  & NA & Ellipsoidal & Coordinate Axes\\
XXX &  $\mathbb{R}^d, K=1$ & $\lambda U D U^T$  & NA & Ellipsoidal & NA\\
\hline
\hline
\end{tabular}}
\caption{List of the \texttt{ssMclust} covariance parameterizations we consider.
In the above, $I$ is the identity matrix; $D$'s are diagonal matrices; and $U$'s
represent matrices of orthonormal eigenvectors.  If ``$k$"
is a
subscript on any symbol then that parameter is allowed to vary
across clusters and, if not, then the parameter must remain
fixed across clusters.  This table is expanded from Table 1 in \cite{fraley2006mclust}.}
\label{tab:mclustModels}
\end{table}

Like in \texttt{Mclust}, the \texttt{ssMclust} framework balances
model fit versus model parsimony.
Like in $\texttt{Mclust}$, we use the Bayesian Information Criterion (BIC)
to assess the quality of the clustering given by the Gaussian Mixture Models with
density structure $f_{\widehat X_v}=\sum_{k=1}^K \pi_k f_{\mu^{(k)},\Sigma^{(k)}}$
over a range of $K$ and various Gaussian parameterizations.
The geometry of the $k$th cluster is determined by the
structure of $\Sigma_k$; see Table \ref{tab:mclustModels} for a comprehensive list
of the covariance structures we consider in \texttt{ssMclust}.
While the more complicated geometric structure allows for a better fit of the data,
this comes at the price of model complexity; i.e., more parameters to estimate.

The BIC penalty employed in \texttt{Mclust} and \texttt{ssMclust} rewards model fit, and it penalizes model complexity.
Given model $M$, the BIC is usually defined as

$$\text{BIC}(M)=2\max_{(\pi,\bmu,\bsigma)\in M}\ell(\pi,\bmu,\bsigma)_{\widehat X,\varphi}-\tau_M\log n,$$

\noindent where $\max_{(\pi,\bmu,\bsigma)\in M}\ell(\pi,\bmu,\bsigma)_{\widehat X,\varphi}$
is the maximized log-likelihood in Eq.~(\ref{loglik}), $\tau_M$ is the number of parameters
estimated in model $M$ (i.e., the number of parameters in $(\pi,\bmu,\bsigma)$
that need to be estimated), and $n$ the number of observed data points.
In the present semi-supervised setting, we propose an adjusted BIC that only penalizes
the model complexity of the unsupervised data points, namely

\begin{align}
\label{eq:BIC'}
\text{BIC}'(M)=2\max_{(\pi,\bmu,\bsigma)\in M}\ell(\pi,\bmu,\bsigma)_{\widehat X,\varphi}-\tau_M\log (n-m).
\end{align}

\noindent If $\lim_{n\rightarrow\infty}m/n=0$, then $|\text{BIC}(M)-\text{BIC}'(M)|=o(1)$,
but even in this setting, empirical evidence suggests the less parsimonious models
allowed by $\text{BIC}'(M)$ provide a better model fit than the more parsimonious $\text{BIC}(M)$.
Intuitively, the complexity introduced by the largely constrained supervised datum
should be lower than that of the unconstrained unsupervised datum, which is reflected in
the modified $\text{BIC}'(M)$; see \cite{yoderdiss}.

The \texttt{ssMclust} algorithm proceeds by maximizing the log-likelihood via the EM
algorithm over a range of models $M\in\mathcal{M}$, and then uses the BIC penalty (\ref{eq:BIC'})
to select the best fitting model, defined via

$$\widehat M = \text{argmax}_{M\in\mathcal{M}} \text{BIC}'(M).$$

\noindent Slightly abusing notation, let $(\hat{\pi},\hat{\bmu},\hat{\bsigma})
:=(\hat{\pi}_{\widehat{M}},\hat{\bmu}_{\widehat{M}},\hat{\bsigma}_{\widehat{M}})$ be
maximum likelihood estimates of $(\pi,\bmu,\bsigma)$ in model $\widehat M$.
The $\mathcal{L}^{EP}$ scheme then nominates the vertices in $A$ via

\begin{align}
\label{eq:espnorder}
{\mathcal L}^{EP}_{G,1}&\in\text{argmax}_{v\in A}\hat\pi_1 f_{\hat\mu^{(1)},\hat\Sigma^{(1)}}(\widehat X_v)\notag;\\
{\mathcal L}^{EP}_{G,2}&\in\text{argmax}_{v\in A\setminus{\mathcal L}^{EP}_{G,1}}\hat\pi_1 f_{\hat\mu^{(1)},\hat\Sigma^{(1)}}(\widehat X_v)\notag;\\
&\vdots\notag\\
{\mathcal L}^{EP}_{G,n-m}&\in\text{argmax}_{v\in A\setminus\left(\cup_{j=1}^{n-m-1}{\mathcal L}^{EP}_{G,j}\right)}\hat\pi_1 f_{\hat\mu^{(1)},\hat\Sigma^{(1)}}(\widehat X_v).
\end{align}

\noindent
Details of the $\mathcal{L}^{EP}$ scheme are summarized in Algorithm \ref{alg:aggClust}.

\begin{algorithm}[t!]
    \DontPrintSemicolon
    \KwIn{
    Graph $G$ on vertices $S\cup A$ (seeds, ambiguous); $n:=|S\cup A|$, $m:=|S|$\newline
    $b\!\restriction_S$ (block assignments of seeds) \newline
    $d$ (embedding dimension)\newline
    $\mathcal{K}$ (maximum number of clusters to consider) \newline
    $\mathcal{M}$ (set of models to consider)
    }
    \KwOut{
    $\mathcal{L}^{EP}$ (nomination scheme)
    }
    $\widehat X\leftarrow$ adjacency spectral embedding of $G$ into $\mathbb{R}^d$;\\
    \ForEach{$M\in\mathcal{M}$}{
    Initialize the class labels using the semi-supervised $K_M-means++$ algorithm;\\
    $\ell_M\leftarrow$ max of complete log-likelihood under model $M$ computed via the EM algorithm;\\
	BIC$'(M)\!\leftarrow 2\ell_M - \tau_M \log(n-m),$ where $\tau_M$ is the number of parameters estimated in~$M$;\\
	}
	$\widehat M \leftarrow \text{argmax}_{M\in\mathcal{M}} \text{BIC}'(M).$\\
	$\mathcal{L}^{EP}\leftarrow$ nomination of the vertices of $A$ according to Eq. (\ref{eq:espnorder}) under model $\widehat M$\\
    \Return{$\mathcal{L}^{EP}$}
    \caption{Extended Spectral Partitioning Vertex Nomination Scheme}
    \label{alg:aggClust}
\end{algorithm}

	In the case of a {\it quasi-seeding}---where $b$ is observed for
vertices in $S_1$ but for vertices in $S\setminus S_1$ it is only observed
that the vertices are not in $V_1$---
the complete data log-likelihood becomes

\begin{align*}
&\ell(\pi,\bmu,\bsigma)_{\widehat X,\varphi} = \sum_{v \in S_1}\log\left(
f_{\mu^{(1)},\Sigma^{(1)}}(\widehat X_v)\right)+  \sum_{k = 2}^K \sum_{v \in S\setminus S_1}  \grr_{  \varphi (v)= k } \log\left(\frac{\pi_k}{1 - \pi_1} f_{\mu^{(k)},\Sigma^{(k)}}(\widehat X_v)\right)\\
&\hspace{45mm} + \sum_{k =1}^K \sum_{v \in A}   \grr_{  \varphi (v)= k }
 \log\left(\pi_k f_{\mu^{(k)},\Sigma^{(k)}}(\widehat X_v)\right),
\end{align*}

\noindent
and Algorithm \ref{alg:aggClust} can be applied with this log-likelihood
in place of Equation~(\ref{loglik}).
The ability of the \texttt{ssMclust} algorithm to seamlessly handle this scenario is a
major advantage over other semi-supervised clustering techniques (e.g., logistic regression, random forest, etc.).

\section{Experimental results}
\label{sec:data}

In this section, we demonstrate the effectiveness (in the sense of precision) and scalability
of our vertex nomination schemes, the canonical sampling vertex nomination
scheme $\mathcal{L}^{CS}$ and the extended spectral partitioning vertex
nomination scheme $\mathcal{L}^{EP}$, on both real and synthetic data.
As mentioned in Section \ref{S:intro}, the canonical vertex nomination scheme
$\mathcal{L}^C$ is optimally effective (in the sense of precision) but does not scale, and the spectral partitioning
vertex nomination scheme $\mathcal{L}^P$ scales well but is not nearly as effective
as $\mathcal{L}^{C}$ on small to medium scale networks. (Indeed, $\mathcal{L}^P$ obtains nearly
chance performance on small graphs). We illustrate in this section that $\mathcal{L}^{CS}$ and
$\mathcal{L}^{EP}$ both scale and are very effective at multiple scales, markedly improving over their forerunners.

Each example in this section consists of $nMC$ Monte Carlo replicates, for some
preselected positive integer $nMC$; that is, we obtain $nMC$ realizations of the
underlying experiment, thus obtaining
$nMC$ nomination lists---for each of the vertex nomination schemes that are compared.
For each vertex nomination scheme, the mean (average) of the $nMC$ average precisions obtained
will be referred to as the {\it empirical mean average precision} under the vertex nomination scheme.
For each vertex nomination scheme and each nomination list position $i$,
the fraction of the $nMC$ nomination lists
in which the $i$th list-position (vertex) was truly in $V_1$ is the {\it empirical probability that nomination
list position $i$ is in $V_1$} under the vertex nomination scheme. All of the figures
in this section consist of plotting the empirical probabilities of nomination lists' position
being in $V_1$ (on the $y$-axis) against the respective position in the nomination list (on the $x$-axis).

Note that we distinguish $nMC$, defined above, from $nMCMC$, which will denote the number of
Markov chain Monte Carlo steps used in computing $\mathcal{L}^{CS}$; unless otherwise specified, we
use $nMCMC/2$ steps for burn-in, and the other $nMCMC/2$ steps for actual sampling.

\subsection{Simulation experiments}
\label{S:sims}

In this subsection, Section \ref{S:sims}, we perform simulation experiments for a stochastic block model at three scales:
the underlying model used here is
$G\sim\text{SBM}\left(3,\vec n,b,\Lambda_\alpha \right)$
                 where

                 $$\Lambda_\alpha:=\alpha\begin{bmatrix} 0.5&0.3&0.4\\
                 0.3&0.8&0.6\\
                 0.4&0.6& 0.3\end{bmatrix}+(1-\alpha)\begin{bmatrix} 0.5&0.5&0.5\\
                 0.5&0.5&0.5\\
                 0.5&0.5& 0.5\end{bmatrix},$$

\noindent for $\alpha\in[0,1]$.
We consider three experimental scales, summarized below in Table~\ref{tab:scales}.
\begin{table}[h!]
\center
\begin{tabular}{cccccc}
   \toprule
  \multicolumn{2}{c}{\bf Scale of Experiment} &$\boldsymbol{\vec m}$&$\boldsymbol{\vec n-\vec m}$&$|$\bf A$|$&$\boldsymbol{\alpha}$\\ \midrule 
\multirow{3}{*}{Small scale}&\mbox{small-small-scale} & $[4,0,0]$&
 $[4,3,3]$& 10&$1$ \\ 
&\mbox{medium-small-scale} &  $[4,0,0]$&
 $[7,4,4]$& 15&$1$ \\ 
&\mbox{large-small-scale} & $[4,0,0]$&
 $[8,5,4]$& 17&$1$ \\  \midrule
 \multicolumn{2}{c}{Medium scale}&$[20,0,0]$&
 $[200,150,150]$& 500&$0.3$\\  
 \multicolumn{2}{c}{Large scale}&$[40,0,0]$&
 $[4000,3000,3000]$& 10000&$0.13$\\  \bottomrule
\end{tabular}
\caption{Experimental parameters for the stochastic block model simulations.}\label{tab:scales}
\end{table}

 The parameter $\alpha$ allows us to control how stochastically differentiated
the blocks are from one another; indeed, as $\alpha$ decreases the blocks become more
stochastically homogeneous and, when $\alpha=0$, there is effectively only one block
(the graph is Erd\H os-R\'enyi). Note that the block of interest, $V_1$, is of intermediate
density; less densely intraconnected than $V_2$ and more than $V_3$. The true model
parameters---$K,\vec n,\Lambda$---are used when implementing
$\mathcal{L}^{C}$, $\mathcal{L}^{CS}$, (as well as $\mathcal{L}^{ML}$---the likelihood
maximization vertex nomination scheme introduced in \cite{fishkind2015vertex}, when relevant),
the true model parameter $K=3$ is used when implementing $\mathcal{L}^P$ (i.e.~$3$-means clustering is applied),
and  $\mathcal{K}=4$ is used in Algorithm \ref{alg:aggClust} when implementing  $\mathcal{L}^{EP}$.

We first compare the effectiveness and runtime of $\mathcal{L}^C$ and $\mathcal{L}^{CS}$ in the small scale regime,
which is the only scale on which $\mathcal{L}^C$ can be feasibly implemented.
In implementing $\mathcal{L}^{CS}$ we used $nMCMC=10000$, with $nMCMC/2=5000$ of these steps discarded as a burn-in.
Results from the $nMC=10000$ experiment realizations are summarized in Table \ref{tab:runtimeCCS} and Figure~\ref{fig:CvsCS}.
\begin{figure*}  
  \centering \vspace{-10mm}
  \subfloat[][{\bf Small-small}]{\includegraphics[width=0.45\textwidth]{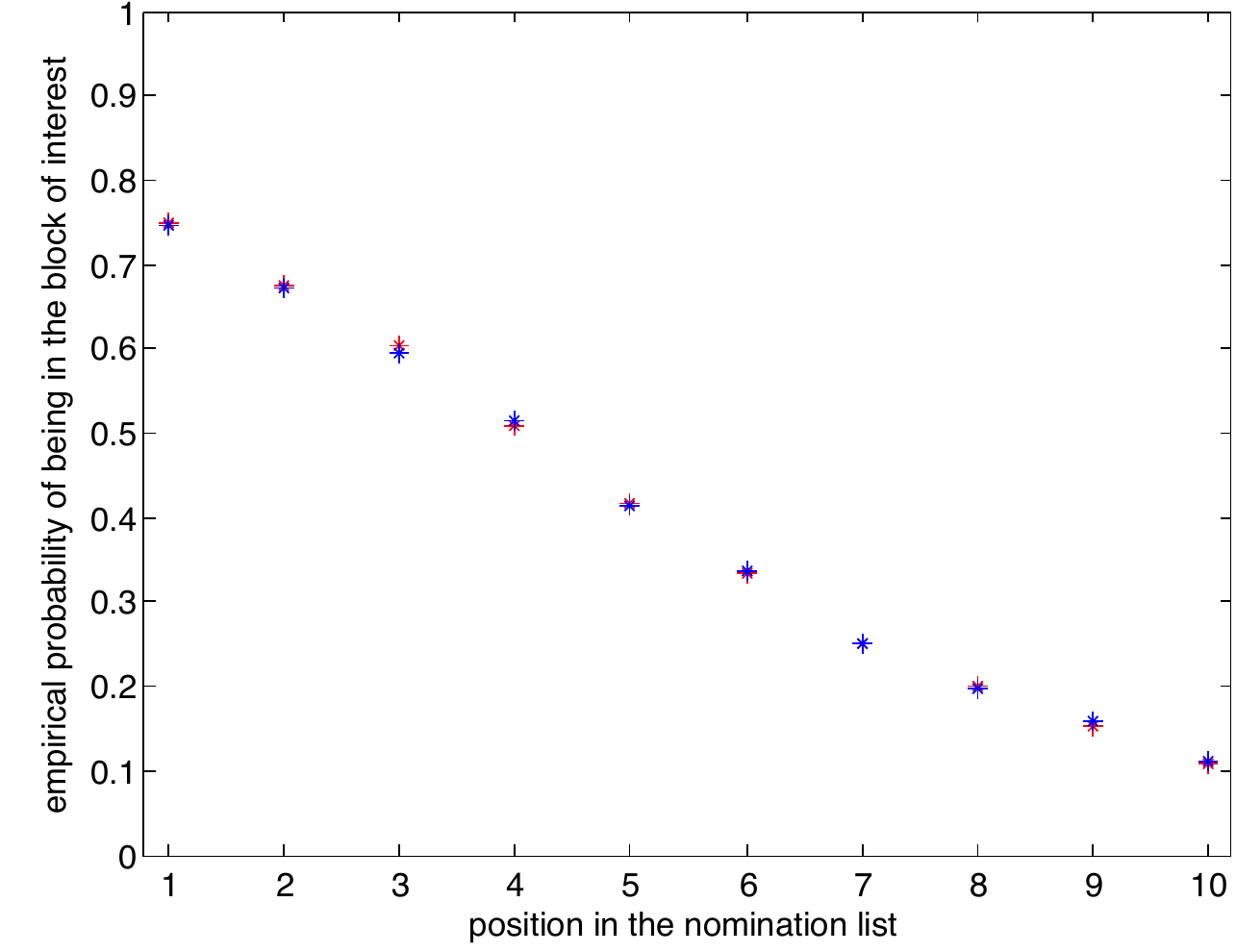}}\hspace{5mm}
  \subfloat[][{\bf Medium-small} ]{\includegraphics[width=0.45\textwidth]{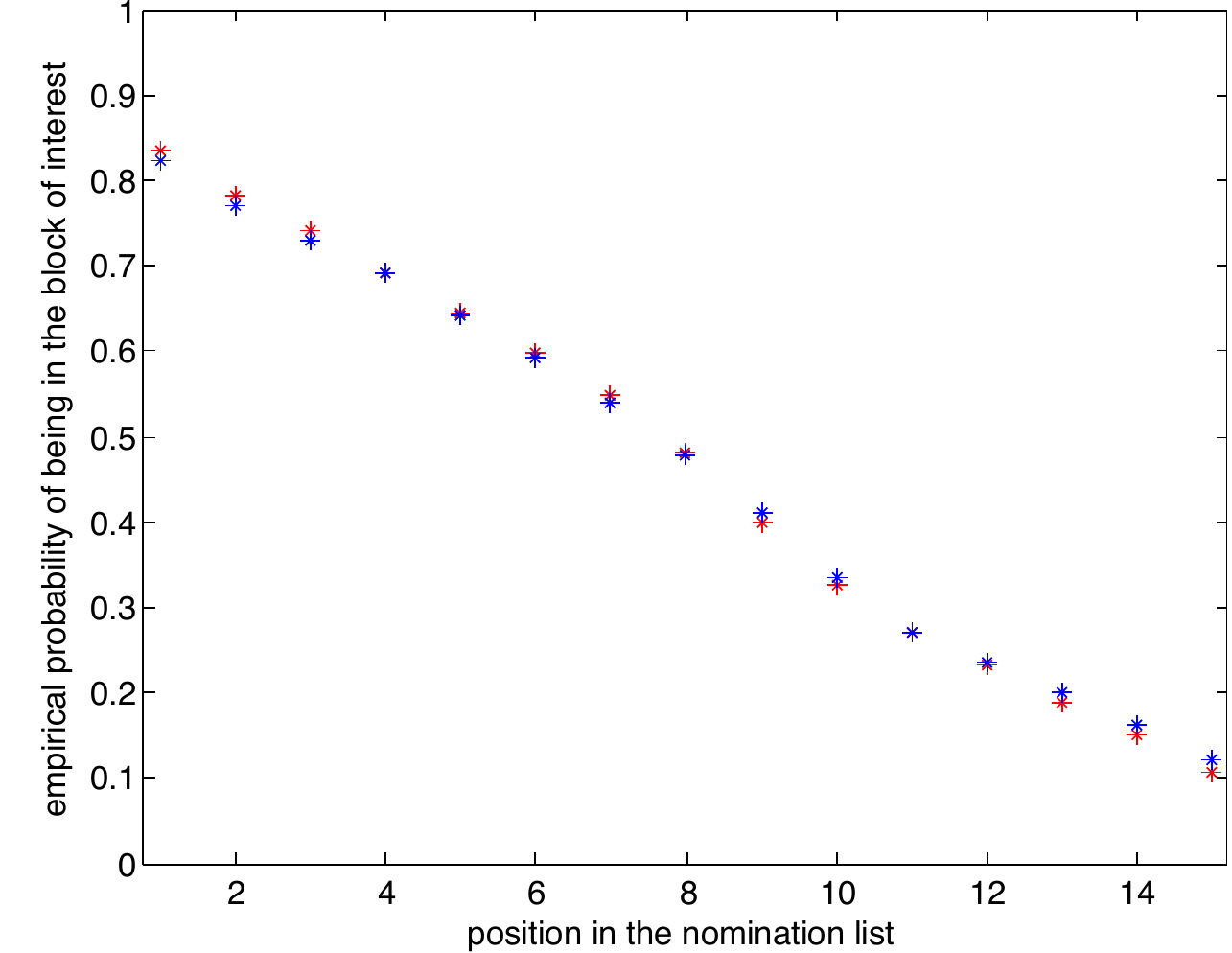}}\\
    \subfloat[][{\bf Large-small}  ]{\includegraphics[width=0.45\textwidth]{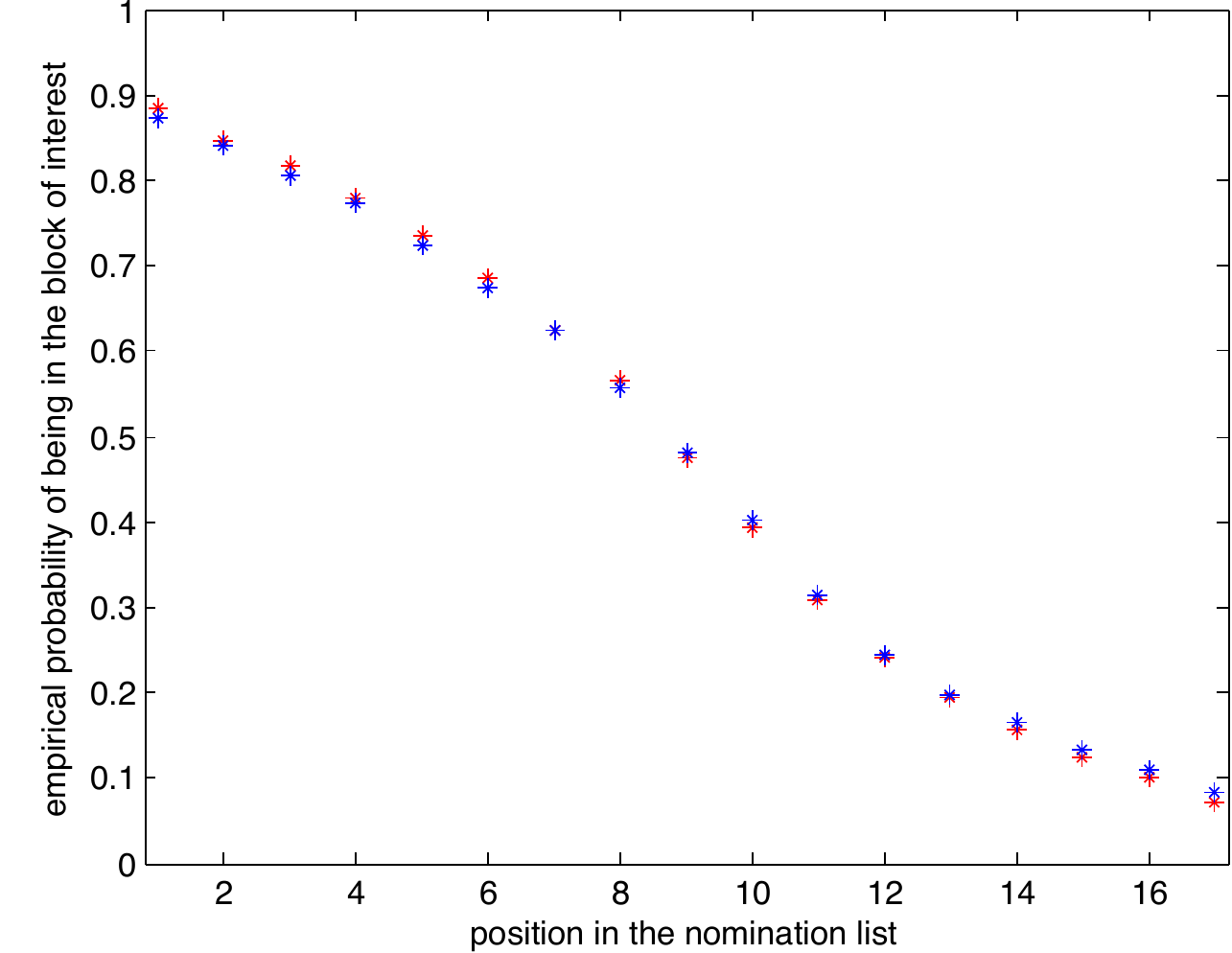}}
  \caption{{\bf Small scale simulations.}
  Empirical probability of being in $V_1$ ($y$-axis) plotted against the respective position in the nomination list ($x$-axis) for $\mathcal{L}^C$ (red) and $\mathcal{L}^{CS}$ (blue).
Here $nMC = 10000$, and for $\mathcal{L}^{CS}$ we use $nMCMC = 10000$; with $nMCMC/2=5000$ steps used for burn-in. (Note that some red asterisks in
these figures are
partially or nearly completely obscured by blue asterisks on top of them.)}
\label{fig:CvsCS}
\end{figure*}
\begin{table} 
\center
\begin{tabular}{lcccccc}
   \toprule
  \bf{Scale of experiment}&$|\mathbf{A}|$ &\multicolumn{2}{c}{\bf Avg. Running Time (in sec.)} &&\multicolumn{2}{c}{\bf MAP} \\ \cmidrule{3-4} \cmidrule{6-7}
&& $\mathcal{L}^C$ & $\mathcal{L}^{CS}$ && $\mathcal{L}^C$ & $\mathcal{L}^{CS}$\\ \midrule
\mbox{small-small-scale} & 10 & $1.12$ & $.0335$ && .6934 & .6901 \\ 
\mbox{medium-small-scale}& 15 & $128$  & $.0453$ && .7632 & .7530 \\ 
\mbox{large-small-scale} & 17 & $871$  & $.0489$ && .8182 & .8086 \\  \bottomrule
\end{tabular}
\caption{{\bf Small scale experiment.}  Comparing $\mathcal{L}^C$ and $\mathcal{L}^{CS}$ by average runtime, empirical MAP.}\label{tab:runtimeCCS}
\end{table}

Observe that $\mathcal{L}^{CS}$ obtains the optimal effectiveness of $\mathcal{L}^{C}$
while running orders of magnitude faster than $\mathcal{L}^{C}$; note that
the running time of $\mathcal{L}^{CS}$ is relatively constant at each
of the three small scale experiments while, empirically, the running of time $\mathcal{L}^{C}$
scales at rate about $2.6^{|A|}$; see Table~\ref{tab:runtimeCCS}.
Indeed, $\mathcal{L}^{CS}$ can be efficiently implemented on graphs with hundreds of thousands
of vertices while $\mathcal{L}^C$ cannot be practically implemented on graphs with more than a few tens of vertices.
At this small scale, we did not include the spectral-based vertex nomination schemes $\mathcal{L}^{EP}$ and
$\mathcal{L}^{P}$, because they are essentially ineffective at this small scale, since the eigenvectors
contain almost no signal, as noted in \cite{fishkind2015vertex}.
\\

Next we move to the medium scale and large scale experiments, with
stochastic block model parameters as given in Table \ref{tab:scales}.
We did $nMC=100$ experiment replicates for each of the vertex nomination schemes;
$\mathcal{L}^{CS}$, $\mathcal{L}^{EP}$, $\mathcal{L}^{P}$, and we also included
the likelihood maximization vertex  nomination scheme $\mathcal{L}^{ML}$ introduced in \cite{fishkind2015vertex},
since it was demonstrated in \cite{fishkind2015vertex,VN3} that $\mathcal{L}^{ML}$
obtains state-of-the-art effectiveness when implementable (i.e., for graphs of order
at most a few thousand vertices). The canonical sampling vertex nomination scheme
$\mathcal{L}^{CS}$ was performed in two ways; once with $nMCMC=100000$, and once
with $nMCMC$ chosen to be such that the runtime of $\mathcal{L}^{CS}$ is equal
to the runtime of $\mathcal{L}^{EP}$. The canonical vertex nomination scheme $\mathcal{L}^{C}$
was not performed in the medium scale and large scale, nor the likelihood
maximization vertex nomination scheme  $\mathcal{L}^{ML}$ at the large
scale, because they are not practical to compute at these scales.
The results of these simulations are summarized
in Table \ref{tab:runtimePEP} and in  Figure \ref{fig:SvsES}.
\begin{figure*}  
  \centering
  \vspace{-20mm}
  \subfloat[][{\bf Medium scale:}\\ $\mathcal{L}^{EP}$ (red) versus $\mathcal{L}^{P}$ (grey). ]{\includegraphics[width=0.45\textwidth]{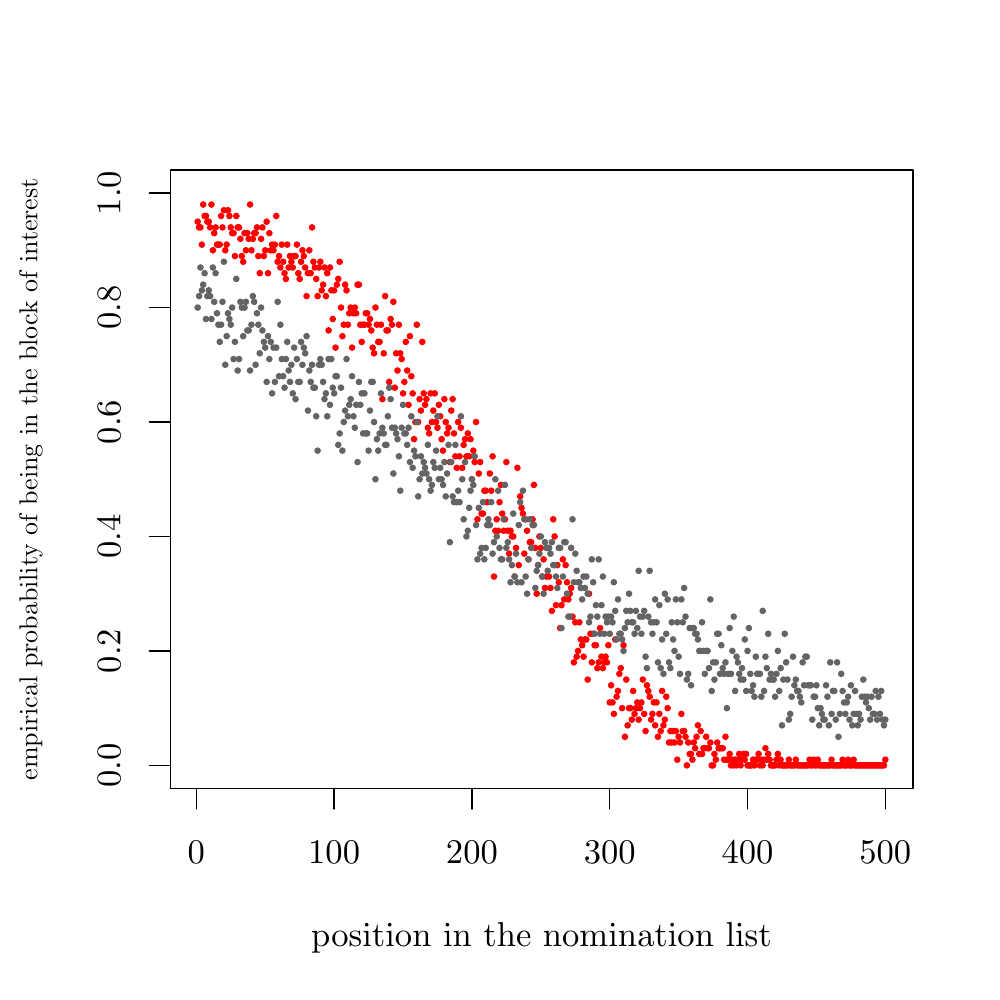}}
  \hspace{5mm}
  \subfloat[][{\bf Large scale:}\\ $\mathcal{L}^{EP}$ (red) versus $\mathcal{L}^{P}$ (grey).  ]{\includegraphics[width=0.45\textwidth]{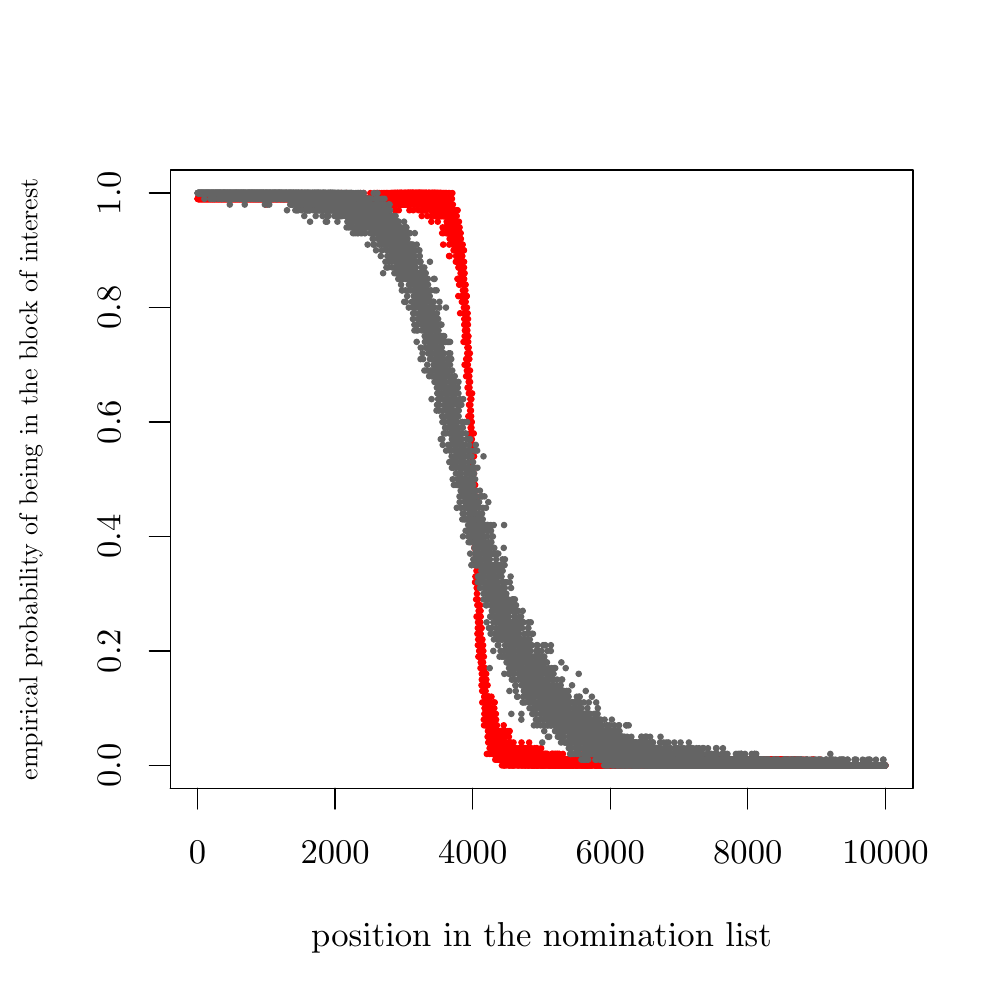}}\\
    \subfloat[][{\bf Medium scale:}\\ $\mathcal{L}^{EP}$~(red); $\mathcal{L}^{CS}$ with $nMCMC$ chosen so that $\mathcal{L}^{EP}$ runtime is same (blue);\\ $\mathcal{L}^{CS}$ with $nMCMC=100000$ (cyan); $\mathcal{L}^{ML}$ (orange).]{\includegraphics[width=0.45\textwidth]{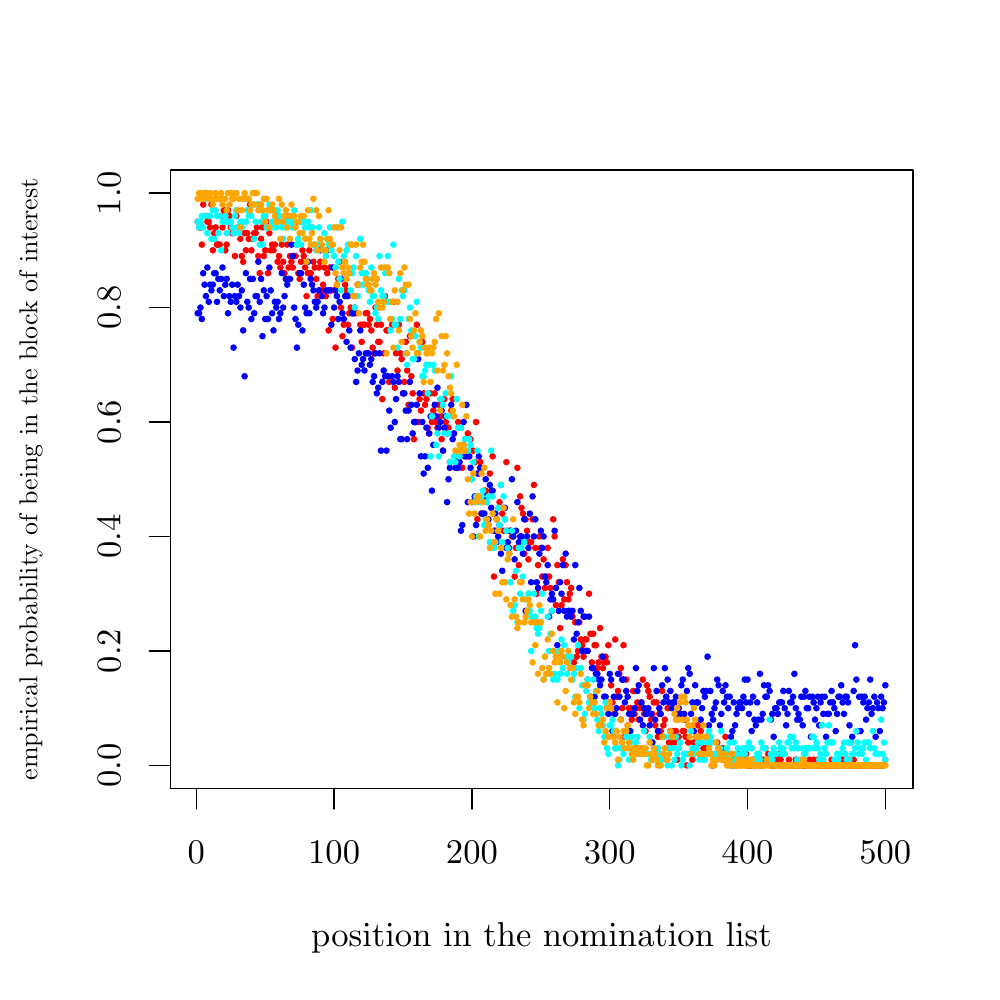}}\hspace{5mm}
  \subfloat[][{\bf Large scale:}\\ $\mathcal{L}^{EP}$ (red); $\mathcal{L}^{CS}$ with $nMCMC$ chosen so that $\mathcal{L}^{EP}$ runtime is same (blue);\\ $\mathcal{L}^{CS}$ with $nMCMC=100000$ (cyan). ]{\includegraphics[width=0.45\textwidth]{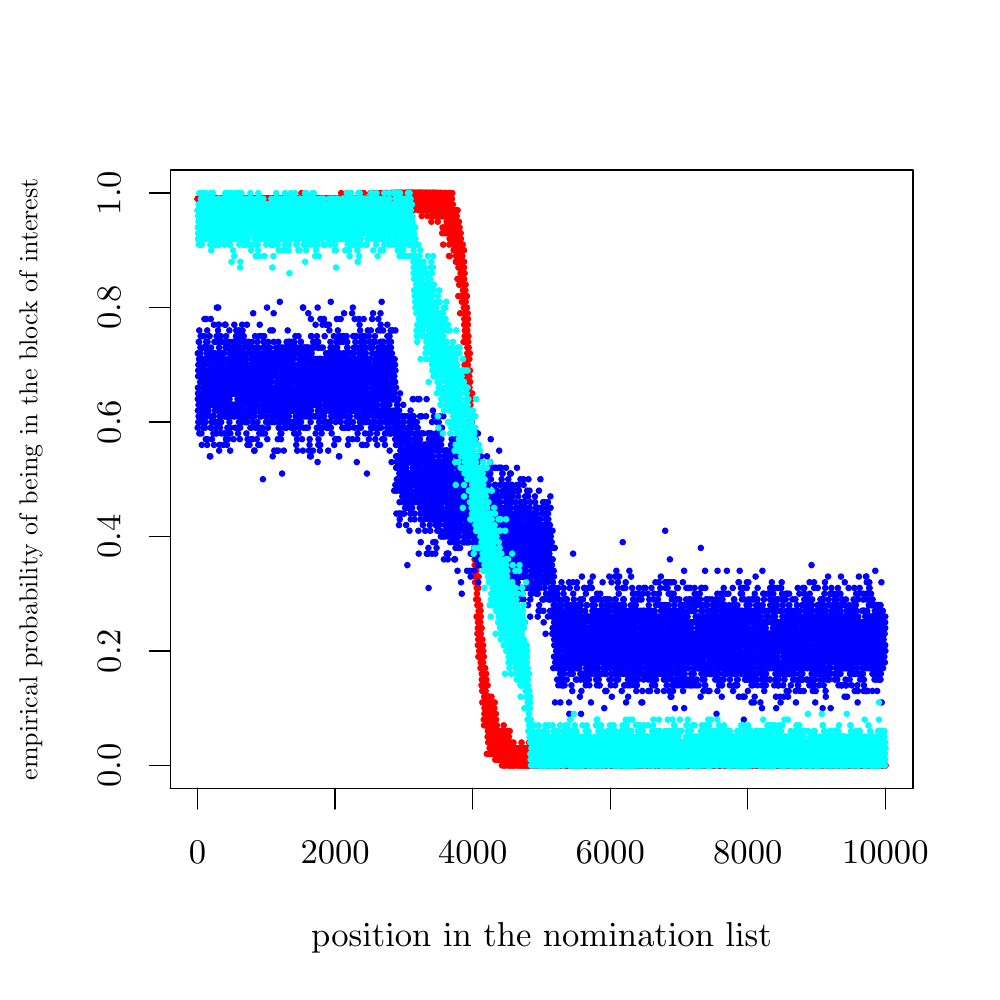}}
  \caption{
  Empirical probability of being in $V_1$ ($y$-axis) plotted against the respective
  position in the nomination list ($x$-axis) for the medium scale (left panels) and
  large scale (right panels) stochastic block model experiments.}
\label{fig:SvsES}
\end{figure*}
\begin{table}  
\centering
\begin{tabular}{lccccc}
\toprule
   & $\mathcal{L}^P$ & $\mathcal{L}^{EP}$ & $\mathcal{L}^{CS}$; $nMCMC$ set to & $\mathcal{L}^{CS}$; $nMCMC$ & $\mathcal{L}^{ML}$\\
   &&& match runtime of $\mathcal{L}^{EP}$ & set to $100000$ & \\
\midrule
  \bf{Scale} & \multicolumn{5}{c}{\bf Running Time (in sec.)} \\
\midrule
\mbox{medium } & ~~0.24 & ~~0.44 & $\longleftarrow$ same & ~~~2.06 & 216.45 \\ 
\mbox{ large } & 19.42  & 19.57 & $\longleftarrow$ same & 112.51 &*\\
\midrule
  \bf{Scale} & \multicolumn{5}{c}{\bf MAP$\pm$2 s.e.}\\
\midrule
\mbox{medium } & $.74\pm.02$ & $.89\pm.02$ & $.80\pm.01$ & $.93\pm.00$ & $.95\pm.00$ \\ 
\mbox{ large } & $.99\pm.02$ & $.99\pm.02$ & $.66\pm.00$ & $.95\pm.00$ &*\\
\bottomrule
\end{tabular}
\caption{{\bf Medium and large scale experiments.}  Comparing $\mathcal{L}^P$, $\mathcal{L}^{EP}$,
$\mathcal{L}^{CS}$ and $\mathcal{L}^{ML}$ by average runtime and empirical MAP. }\label{tab:runtimePEP}
\end{table}

First, observe that in both the medium and the large scale $\mathcal{L}^{EP}$ was more effective than $\mathcal{L}^{P}$,
significantly so in the medium scale regime, with a twofold runtime increase being the cost for this  increase
in effectiveness.
In the adjacency spectral embedding
of a stochastic block model, the within-class variance is, with high probability, of the order
$\frac{\log n}{\sqrt{n}}$; see \cite{perfect}.
Thus, as there are more vertices, the true clusters
become more easily delineated, and the adjacency spectral clustering step of
$\mathcal{L}^{EP}$ and of $\mathcal{L}^{P}$ is dominated in running time by
the embedding step, which is the same for $\mathcal{L}^{EP}$ and $\mathcal{L}^{P}$.
However, in the medium scale regime, where the true clusters are less
easily recovered in the embedding, the more sophisticated clustering procedure utilized
in $\mathcal{L}^{EP}$ is significantly more effective than the $k$-means clustering used in
$\mathcal{L}^{P}$---at the expense of an increase in runtime.

In the medium scale regime, while we see that $\mathcal{L}^{ML}$ is the most effective of
the vertex nomination schemes that we compare, note that
the runtime of $\mathcal{L}^{ML}$ was orders of magnitude  greater then the other
vertex nomination schemes. In fact, $\mathcal{L}^{ML}$ is
not practical to implemented on graphs with more than a few thousand vertices (such as our
large scale experiment), unlike $\mathcal{L}^{CS}$ and $\mathcal{L}^{EP}$.
In both the medium and large scale examples,
we see that $\mathcal{L}^{EP}$ is significantly more effective than $\mathcal{L}^{CS}$ when $\mathcal{L}^{CS}$
is restricted to have the same running time as $\mathcal{L}^{EP}$. However, $\mathcal{L}^{CS}$ will
eventually be more effective than $\mathcal{L}^{EP}$ (and all other vertex nomination schemes other than $\mathcal{L}^C$)
given enough Markov chain Monte Carlo steps.

Indeed, to illustrate the effects of increasing the amount of sampling on $\mathcal{L}^{CS}$,
we repeated the experiment  in both the medium and large scales for $\mathcal{L}^{CS}$ with
values $nMCMC=10^3,\ 10^4,\ 10^5,$ and $10^6$.
The results of $nMC=1000$ realizations are shown in Figure \ref{fig:my}. In the medium scale,
from $nMCMC=10^4$ and up, the increased sampling still improved the effectiveness but
seemed to stabilize towards a limit. In the large scale, continued steady improvement in
effectiveness was seen for the increases in $nMCMC$, until $nMCMC=10^6$ allowed for the near
perfect success in the nomination task.
\begin{figure*}[t!]  
  \centering
  \vspace{-20mm}
  \subfloat[][{\bf Medium scale} ]{\includegraphics[width=0.45\textwidth]{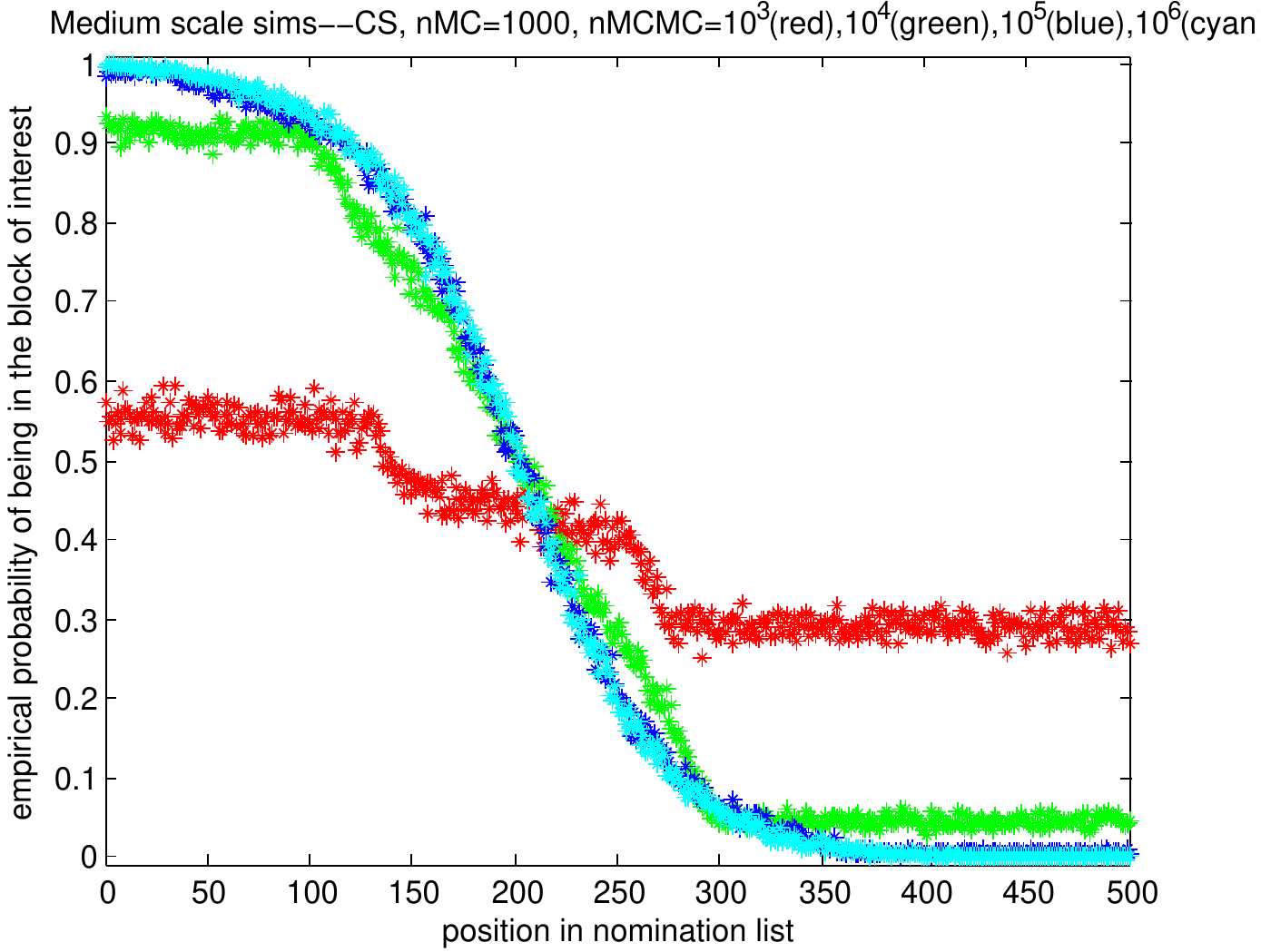}}
  \hspace{5mm}
  \subfloat[][{\bf Large scale} ]{\includegraphics[width=0.45\textwidth]{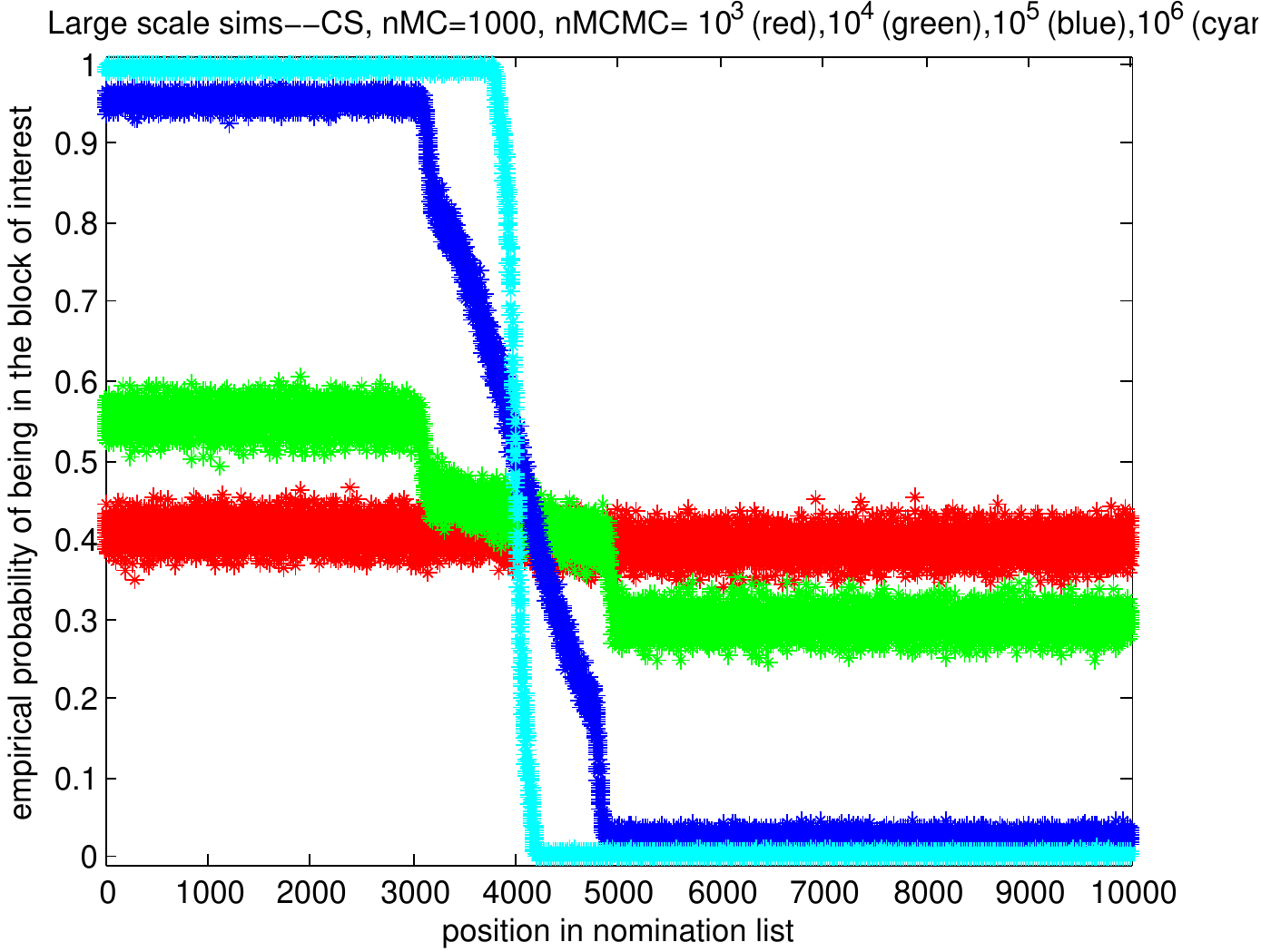}}
  \caption{
  The effect on $\mathcal{L}^{CS}$ of increasing the value of $nMCMC$;
   plots are shown for    $nMCMC=10^3$ (red), $nMCMC=10^4$ (green), $nMCMC=10^5$ (blue), $nMCMC=10^6$ (cyan).}
\label{fig:my}
\end{figure*}

\subsection{More simulation experiments \label{sec:simmore} }

In this subsection, Section \ref{sec:simmore}, we perform more simulation experiments to explore
the tradeoff, for $\mathcal{L}^{CS}$ and $\mathcal{L}^{EP}$, between computational burden and effectiveness (i.e.~precision). We also consider
the effect of embedding dimension on the performance of $\mathcal{L}^{EP}$
since, in practice, the correct value of $d=\textup{rank} \Lambda$ may not be known for use in the implementation of $\mathcal{L}^{EP}$.

In particular, in this subsection, the {\it embedding dimension}
will refer to a positive integer $\eth$ that will replace $d$ everywhere
in the adjacency spectral embedding step
of Section \ref{Sec:spec}
(thus the vertices are embedded into ${\mathbb R}^\eth$ instead of ${\mathbb R}^d$)---and $\eth$ will also replace $d$ onward in the definition of $\mathcal{L}^{EP}$ as given in Section \ref{Sec:ssclust}.
The results in \cite{fishkind2013consistent} imply that the effectiveness
of $\mathcal{L}^{EP}$ should not degrade too much if $\eth>d$, but \cite{fishkind2013consistent} includes an example (beginning of Section 8, see Figure 1) where $\eth<d$ leads to a complete breakdown
in spectral partitioning, with performance almost as bad as chance.
In the setting we experiment with here,
the effectiveness of $\mathcal{L}^{EP}$ will be seen as
relatively robust to overestimation as well as underestimation of $d$.

Here we will use the following parameters: \ \ $K=10$;

$$  \Lambda =
\left [ \begin{array}{rrrrrrrrrr}
.30 & .27 & .24 & .21 & .21 & .21 & .21 & .21 & .21 & .21 \\
.27 & .30 & .27 & .24 & .21 & .21 & .21 & .21 & .21 & .21 \\
.24 & .27 & .30 & .27 & .24 & .21 & .21 & .21 & .21 & .21 \\
.21 & .24 & .27 & .30 & .27 & .24 & .21 & .21 & .21 & .21 \\
.21 & .21 & .24 & .27 & .30 & .27 & .24 & .21 & .21 & .21 \\
.21 & .21 & .21 & .24 & .27 & .30 & .27 & .24 & .21 & .21 \\
.21 & .21 & .21 & .21 & .24 & .27 & .30 & .27 & .24 & .21 \\
.21 & .21 & .21 & .21 & .21 & .24 & .27 & .30 & .27 & .24 \\
.21 & .21 & .21 & .21 & .21 & .21 & .24 & .27 & .30 & .27 \\
.21 & .21 & .21 & .21 & .21 & .21 & .21 & .24 & .27 & .30 \\
\end{array} \right ], \
\vec{n} =
\left [ \begin{array}{r}
100 \\ 100 \\ 100 \\ 100 \\ 100 \\ 100 \\ 100 \\ 100 \\ 100 \\ 100
\end{array} \right ], \
\vec{m} =
\left [ \begin{array}{r}
20 \\ 20 \\ 20 \\ 20 \\ 20 \\ 20 \\ 20 \\ 20 \\ 20 \\ 20
\end{array} \right ] .
$$

These parameters were chosen so that the blocks are stochastically
similar to each other, there are many blocks, and the differences
between the probabilities in $\Lambda$ are mild relative to
the number of vertices involved; all of these factors make the
vertex nomination task quite challenging,
since there is a limited amount of signal present.

For each value of embedding dimension
$\eth=2, 3, 4, 5, 8, 9, 10, 11, 12, 15, 20$
we obtained $nMC=200$
independent
realizations of the random graph with the above parameters, and we
nominated for the block of interest $V_1$ using the extended spectral
partitioning vertex nomination scheme $\mathcal{L}^{EP}$, and we recorded
the mean runtime and the also the
empirical mean average precision. We also used the
canonical sampling vertex nomination scheme $\mathcal{L}^{CS}$
on these realizations, but
chose the number of Markov chain Monte Carlo steps $nMCMC$ so that
the runtime was the same (``equitimed") as the mean
$\mathcal{L}^{EP}$ runtime;
we recorded the empirical mean average precision from this
``equitimed" $\mathcal{L}^{CS}$.
We also used the
canonical sampling vertex nomination scheme $\mathcal{L}^{CS}$
again on these realizations, but now we
allowed the number of Markov Chain Monte Carlo steps $nMCMC$
to be exactly as large as needed to acheive
equal empirical mean average
precision as was achieved by $\mathcal{L}^{EP}$; we recorded the
mean runtime of this ``equiprecise" $\mathcal{L}^{CS}$.
(Because the value of $nMCMC$ was
not known a priori, we fixed the burn-in
for $\mathcal{L}^{CS}$ in this subsection at $T=5000$.)
The results of these experiments are displayed in Table \ref{table:comp}.

 \begin{table}[ht]
 \centering
 \begin{tabular}{lrrrrrrrrrrr}
 \toprule
embedding dimension $\eth$ & 2 & 3 & 4 & 5 & 8 & 9 & 10 & 11 & 12 & 15 & 20   \\
   \midrule
 ${\mathcal L}^{EP}$ MAP & .41 & .53 & .53 & .51 & .49 & .50 & .49 & .49 & .49 & .48 & .47  \\
 ${\mathcal L}^{EP}$ time  & .50 & .60 & .84 & 1.01 & 1.71 & 2.02 & 2.37 & 2.72 & 3.09 & 4.39 & 6.31  \\
${\mathcal L}^{CS}$ equitime MAP & .13 & .16 & .25 & .28 & .33 & .36 & .39 & .41 & .44 & .49 & .56  \\
${\mathcal L}^{CS}$ equiprecise time  & 3.01 & 5.74 & 5.69 & 5.30 & 5.01 & 5.01 & 4.99 & 4.52 & 5.08 & 4.74 & 4.05  \\
    \bottomrule
 \end{tabular}
 \caption{Trade-off of computational burden vs.~precision between
 ${\mathcal L}^{EP}$ and ${\mathcal L}^{CS}$, and also comparison
 across different embedding dimensions. All times in this table are
 the average number of seconds, and all values of MAP are +/- .01.
 The runtimes in the bottom row---${\mathcal L}^{CS}$
 equiprecise time---have standard error ranging from $.13$ to $.48$, most
 are approximately~$.24$. The runtimes in the second
 row---${\mathcal L}^{EP}$ time---have standard error ranging from $.01$ to $.09$, most are approximately $.04$.
 \label{table:comp}}
 \end{table}

Note that when devoting the same computational resources
to $\mathcal{L}^{CS}$ and  $\mathcal{L}^{EP}$, we saw that here,
for smaller values of $\eth$, \ $\mathcal{L}^{EP}$ achieved
higher mean average precision than did $\mathcal{L}^{CS}$ and,
for larger values of $\eth$, \   $\mathcal{L}^{CS}$ achieved
higher mean average precision than did $\mathcal{L}^{EP}$.
This is because $\mathcal{L}^{EP}$ took longer and longer to
run in more dimensions, and the increased sampling time
allowed $\mathcal{L}^{CS}$ to pull ahead in precision.
Indeed, the mean average precision of $\mathcal{L}^{EP}$
is terminal, in contrast to $\mathcal{L}^{CS}$, for which
longer and longer sampling times will increase its mean average
precision as long as patience allows---and,
in the limit, to the highest attainable mean average precision.

Also note that the performance of $\mathcal{L}^{EP}$ here was relatively
robust for incorrect embedding dimension ($\eth$ being greater
or lesser then $d$). Although \cite{fishkind2013consistent}
highlights by example the dangers of underestimating $d$, this example
illustrates that such underestimation can be benign.
In particular, $\eth=3,4$ led to
somewhat better performance than the correct value $\eth=d=10$.
This can be explained by the decay in the eigenvalues
of $\Lambda$; here the eigenvalues of $\Lambda$ are
$2.3465$,\ $0.2197$,\ $0.1745$,\ $0.1112$,\
$0.0648$,\ $0.0300$,\ $0.0235$,\ $0.0178$,\ $0.0064$,\ $0.0056$. After
the first four greatest eigenvalues, the rest are small enough to
cause $\Lambda$ to produce behavior similar to that which
a lower rank matrix would produce. Rigorous analysis
of the optimal embedding dimension is beyond the scope of this
present paper; see \cite{elbow} for principled
methodology.

\subsection{Real data example: A human connectome \label{sec:real}}

In this subsection, Section  \ref{sec:real}, we consider a real-data example; a human connectome. This is
a graph with vertices corresponding to locations in a human
brain and edges which reflect functional adjacency. The block structure
that we consider isn't ostensibly reflective of an actual stochastic block model. Indeed,
the vagarities of such real data gives us no reason to expect that there is
precisely an underlying probabilistic block uniformity. Nonetheless, employing
a stochastic block model as an approximation seems to be a plausibly useful
approach. In fact, we will see that all of the important operational
observations of this article do indeed occur here.
Specifically,
on this large graph, where $L^{ML}$ and $L^{C}$ schemes are not practical to implement,
we will see that the nomination schemes introduced in this article scale very well, and
we will see here that the extended spectral partitioning vertex nomination scheme is significantly more
effective than the (original) spectral partitioning vertex nomination scheme,
and the canonical sampling vertex nomination scheme is more effective than
both---when enough computation is performed.

The human connectome (brain graph) that we use here comes from the very
recent paper \cite{Kiar2017}; the particular connectome that we employ is actually
one level of a multiscale hierarchy provided there, and this hierarchy is sure to be a
rich object of study in future work. Our graph was obtained as follows.
Two diffusion MRI (dMRI) and two structural MRI (sMRI) scans were done on
an individual, collected over two sessions \cite{corr}. Graphs were estimated using the
NDMG \cite{corr} pipeline. The dMRI scans were pre-processed for eddy currents using
FSL's \texttt{eddy-correct} \cite{eddy}. FSL's ``standard" linear registration
pipeline was used to register the sMRI and dMRI images to the
MNI152 atlas \cite{fsl1,fsl2,fsl3,mni152}. A tensor model was fit
using DiPy \cite{dipy} to obtain an estimated tensor at each voxel.
A deterministic tractography algorithm was applied using DiPy's
EuDX \cite{dipy, eudx} to obtain a fiber streamline from each voxel.
Graphs were formed by contracting fiber streamlines into sub-regions depending on
spatial \cite{glocal} proximity or neuro-anatomical
\cite{aal, desikan, harvardoxford, talairach, jhu, glasser, pvt, slab907, slab1068} similarity;
we used neuro-anatomical similarity.

\begin{figure}[ht!]
  \centering
  \includegraphics[width=0.7\textwidth]{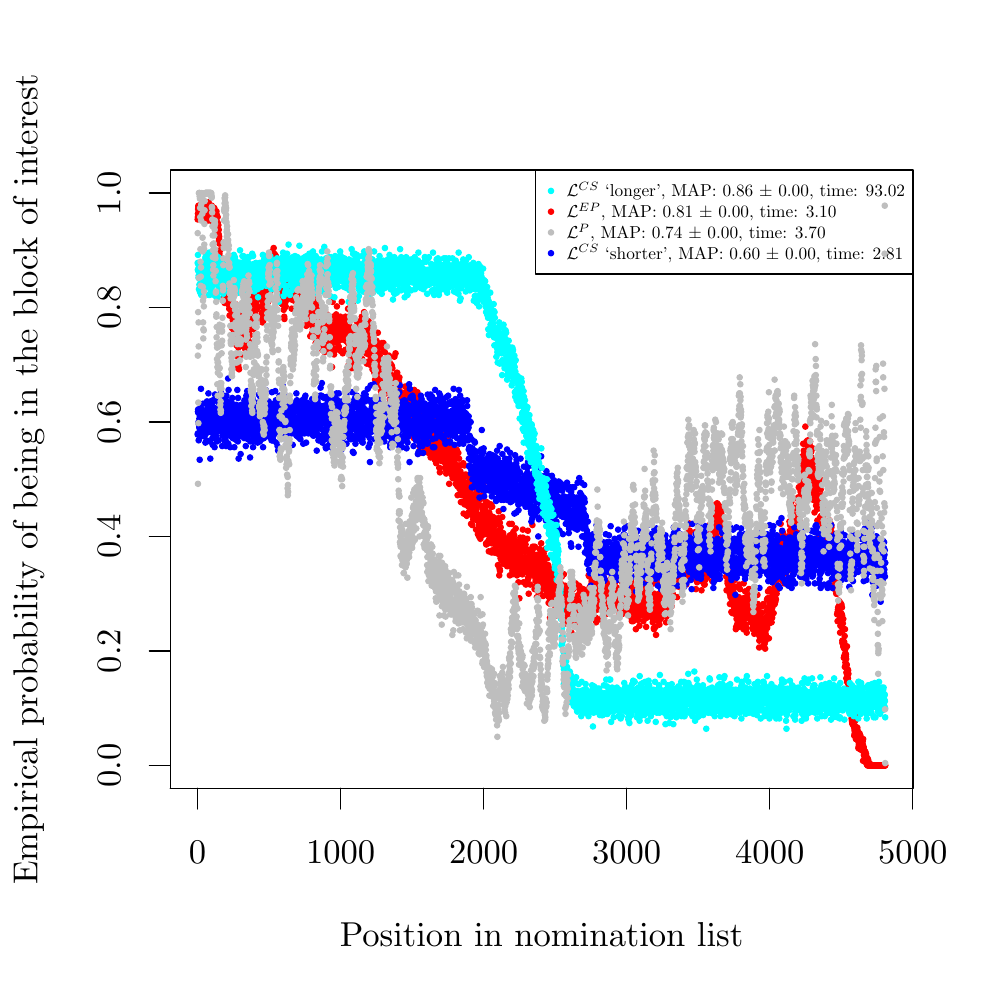}
  \caption{For the connectome real-data experiments,
  comparing the effectiveness of ${\mathcal L}^{P}$ (gray),
  ${\mathcal L}^{EP}$ (red), ``shorter" ${\mathcal L}^{CS}$ (blue),
  and ``longer" ${\mathcal L}^{CS}$ (cyan). }
\label{fig:JYYP}
\end{figure}

We consider a three block SBM model for this data; $V_1$ are the regions corresponding to the right
hemisphere, $V_2$ are the regions corresponding to the left hemisphere,
and $V_3$ are regions that are not characterized. In particular,
$n_1=2807$, $n_2=2780$, and $n_3=271$. The number of seeds we considered were
$m_1=500$, $m_2=500$, $m_3=50$, respectively; in each of $nMC=500$
experiment replicates, we independently
discrete-uniformly selected the seeds from the blocks, and
constructed a nomination
list for the remaining $4808$ ambiguous vertices using
each of vertex nomination schemes
${\mathcal L}^{P}$, ${\mathcal L}^{EP}$, ``shorter" ${\mathcal L}^{CS}$,
and ``longer" ${\mathcal L}^{CS}$.
``Longer" ${\mathcal L}^{CS}$ used $nMCMC=100000$ and ``shorter" ${\mathcal L}^{CS}$ used
$nMCMC=3000$, the latter value chosen so that
 ${\mathcal L}^{CS}$  runtime was approximately
the same as the runtime of ${\mathcal L}^{EP}$. Both
${\mathcal L}^{P}$ and ${\mathcal L}^{EP}$ used embedding
dimension $d=6$ (since this was the first elbow in the scree plot
as determined through the algorithm of Zhu and Ghodsi \cite{zhu2006automatic});
${\mathcal L}^{P}$ used $1000$ k-means restarts, and
${\mathcal L}^{EP}$ considered the
 `EEV', `EEE', and `EII' covariance structures in Table \ref{tab:mclustModels}, and
 ${\mathcal K}=3$ number of clusters. For each of
``shorter" ${\mathcal L}^{CS}$ and ``longer" ${\mathcal L}^{CS}$,
the value of $\Lambda$ was estimated from population densities, and
half of $nMCMC$ steps were burn-in.

\begin{table}[h!]
\center
\begin{tabular}{lcc}
   \toprule
  \bf{Nomination scheme}& \bf{MAP} &  \bf{ Avg. Running Time} \\ \midrule
${\mathcal L}^{CS}$ \mbox{ ``longer" }  & .86 &  93.02 \mbox{sec.}  \\ 
${\mathcal L}^{EP}$  &  .81 &  3.10 \mbox{sec.}  \\ 
${\mathcal L}^{P}$  & .74  & 3.70 \mbox{sec.}  \\ 
${\mathcal L}^{CS}$ \mbox{ ``shorter" }  & .60 &  2.81 \mbox{sec.}  \\ \bottomrule
\end{tabular}
\caption{Comparison of MAP and runtimes for vertex nomination schemes on the
connectome.}\label{tab:connectome}
\end{table}
The results of these experiments are summarized in Table~\ref{tab:connectome}
and Figure~\ref{fig:JYYP}.
In particular, note that ${\mathcal L}^{EP}$ was
substantially more effective than ${\mathcal L}^{P}$, although their runtimes were about
the same. Also note that when ${\mathcal L}^{CS}$ was limited in runtime to the order of
runtime for ${\mathcal L}^{EP}$, it was not competitive in terms of effectiveness but, with increased runtime,
${\mathcal L}^{CS}$ did eventually overtake all of the other vertex nomination schemes in terms
of effectiveness. On a graph of this order, having approximately $5000$ ambiguous vertices,
the  likelihood maximization vertex nomination scheme $\mathcal{L}^{ML}$ and the
canonical vertex nomination scheme $\mathcal{L}^C$ were not tractable. Indeed, these experiments
highlight the scalability and effectiveness of the vertex nomination schemes
${\mathcal L}^{CS}$ and ${\mathcal L}^{EP}$ introduced in this paper.

\section{Summary and future directions}

In summary, for a vertex nomination instance,
the optimally precise vertex nomination
scheme---the canonical nomination scheme $\mathcal{L}^C$---is only
practical for the smallest, toy problems. For larger instances,
the likelihood maximization nomination scheme $\mathcal{L}^{ML}$
should be used, until the size of the problem is too big for
this to be practical, which may be on the order of a
thousand or so vertices. For larger instances, the
extended spectral partitioning $\mathcal{L}^{EP}$ and the
canonical sampling  $\mathcal{L}^{CS}$ vertex nomination schemes
(introduced in this paper) should be used;
the former can be the better choice when computational
resources are more limited and less is known about the
model parameters, and the latter can be the
better choice when there is more knowledge of the model
parameters and there are greater computational resources.

Concurrent work in vertex nomination has tackled the nomination problem in a slightly modified setting, considering a pair of networks and using vertices of interest in one network to nominate potential vertices of interest in the second network \cite{HP,LKP}.  In this paired graph setting, the concept of nomination consistency is established for general network models (and for a general notion of ``vertices of interest") in \cite{LKP,ag1}, and the surprising fact that universally consistent vertex nomination schemes do not exist is established in \cite{LKP}.  In the present, single network setting, this points to a direction for future research: Generalizing the concept of vertices of interest beyond community membership, and establishing the statistical framework for vertex nomination consistency in the setting where more general vertex covariates delineate ``interesting" versus ``non-interesting" vertices.  \\

\noindent {\bf Acknowledgements:} The authors are grateful to the
referees and editors for very useful feedback that greatly enhanced
this paper.
Support in part provided by the Johns Hopkins University Human Language Technology CoE,
the DARPA SIMPLEX program through contract N66001-15-C-4041,
the DARPA D3M program through contract FA8750-17-2-0112,
and the Acheson J. Duncan Fund for the Advancement of Research in Statistics at Johns Hopkins University. The work of authors JY, LC, HP was undertaken while graduate students at Johns Hopkins University.

\bibliographystyle{plainnat}
\bibliography{thebib}

\end{document}